\documentclass[preprint,12pt,authoryear]{elsarticle}

\usepackage{amssymb}

\usepackage[hidelinks]{hyperref}
\usepackage{booktabs}
\usepackage{siunitx}
\usepackage{xcolor}
\usepackage{array}
\usepackage{longtable}
\usepackage{caption}
\usepackage{rotating}
\usepackage{pdflscape}
\usepackage{csquotes}

\journal{Expert Systems with Applications}

\begin{document}

\begin{frontmatter}



\title{CareerBERT: Matching Resumes to ESCO Jobs in a Shared Embedding Space for Generic Job Recommendations}


\author[inst1]{Julian Rosenberger\corref{cor1}}
\ead{julian.rosenberger@ur.de}
\author[inst2]{Lukas Wolfrum}
\ead{lukas.wolfrum@fau.de}
\author[inst2]{Sven Weinzierl}
\ead{sven.weinzierl@fau.de}
\author[inst1]{Mathias Kraus}
\ead{mathias.kraus@ur.de}
\author[inst3]{Patrick Zschech}
\ead{patrick.zschech@tu-dresden.de}

\cortext[cor1]{Corresponding author}

\affiliation[inst1]{
            organization={Universität Regensburg},
            addressline={Bajuwarenstraße 4}, 
            city={Regensburg},
            postcode={93053},
            country={Germany}
}

\affiliation[inst2]{
            organization={Friedrich-Alexander-Universität Erlangen-Nürnberg},
            addressline={Lange Gasse 20}, 
            city={Nürnberg},
            postcode={90403},
            country={Germany}
}

\affiliation[inst3]{organization={TU Dresden},
            addressline={Helmholtzstraße 10}, 
            city={Dresden},
            postcode={01069}, 
            country={Germany}
}

\begin{abstract}
The rapidly evolving labor market, driven by technological advancements and economic shifts, presents significant challenges for traditional job matching and consultation services. In response, we introduce an advanced support tool for career counselors and job seekers based on CareerBERT, a novel approach that leverages the power of unstructured textual data sources, such as resumes, to provide more accurate and comprehensive job recommendations. In contrast to previous approaches that primarily focus on job recommendations based on a fixed set of concrete job advertisements, our approach involves the creation of a corpus that combines data from the European Skills, Competences, and Occupations (ESCO) taxonomy and EURopean Employment Services (EURES) job advertisements, ensuring an up-to-date and well-defined representation of general job titles in the labor market. Our two-step evaluation approach, consisting of an application-grounded evaluation using EURES job advertisements and a human-grounded evaluation using real-world resumes and Human Resources (HR) expert feedback, provides a comprehensive assessment of CareerBERT's performance. Our experimental results demonstrate that CareerBERT outperforms both traditional and state-of-the-art embedding approaches while showing robust effectiveness in human expert evaluations. These results confirm the effectiveness of CareerBERT in supporting career consultants by generating relevant job recommendations based on resumes, ultimately enhancing the efficiency of job consultations and expanding the perspectives of job seekers. This research contributes to the field of NLP and job recommendation systems, offering valuable insights for both researchers and practitioners in the domain of career consulting and job matching.
\end{abstract}


\begin{highlights}
\item CareerBERT: modular expert system combining ESCO taxonomy with EURES jobs for standardized matching
\item Comprehensive evaluation against traditional systems, transformers and LLMs shows superior performance
\item Modular architecture enables practical deployment while accommodating future language model advances
\item Extensive validation through application metrics and human expert assessment demonstrates reliability
\end{highlights}


\end{frontmatter}

\noindent{\small \copyright~2025. This manuscript version is made available under the CC BY 4.0 license \url{https://creativecommons.org/licenses/by/4.0/}. The published version is available here: \url{https://doi.org/10.1016/j.eswa.2025.127043}.}

\vspace{1em}

\section{Introduction}
\label{sec:intro}

In an era of rapid technological advancements and shifting economic landscapes, the labor market is undergoing profound transformations. Digital innovations, the rise of remote work, automation, and the expansion of the gig economy are fundamentally altering job roles and employment dynamics \citep{teigland2019digital}. This evolution is dual-faceted, fostering the need for new skills while simultaneously displacing a significant portion of the workforce as roles evolve or become obsolete \citep{autor2015there}. The \cite{wef2023jobs} estimates that nearly half of the workforce will require retraining for new roles within the next five years to remain relevant, highlighting the urgency of addressing these changes.

Despite the pressing need for effective job matching and consultation services, traditional job agencies and career services are struggling to adapt to these dynamic changes. Outdated frameworks and a reliance on structured data inputs hinder their ability to match the pace of market developments and capture the rapidly diversifying job roles and trends \citep{musset2018working, lavi2021consultantbert, bhatia2019end}. This misalignment emphasizes the need for more robust support systems that can leverage unstructured data sources, such as resumes, which offer rich information about a candidate's experience, skills, and potential \citep{organisation2019oecd}.

Artificial Intelligence (AI) has significantly transformed the way businesses process, analyze, and utilize data, moving beyond traditional statistical methods to leverage advanced machine learning algorithms and predictive analytics \citep{campbell2020data, feuerriegel2024generative}. AI-driven technologies enable businesses to extract deeper insights from vast and complex datasets, automate decision-making processes, and enhance operational efficiency across various domains such as customer service, supply chain management, and marketing. These advancements have not only increased the speed and accuracy of data analysis but also empowered businesses to make more informed, data-driven decisions. The emerging field of Natural Language Processing (NLP) presents a promising avenue for exploiting unstructured data sources, such as text documents, emails, and social media posts, due to its ability to interpret, analyze, and extract meaningful insights from human language, which is often complex and context-dependent \citep{frankel2022disclosure, borchert2024industry}. Recent advancements in semantic embeddings have further enhanced NLP’s potential by enabling more accurate representation and understanding of textual data, transforming applications like job recommendations. This aligns with broader trends in generative artificial intelligence, which are shaping labor productivity and job displacement \citep{lazaroiu2023generative} and driving innovations in extended reality, enterprise management algorithms, and Internet of Things (IoT) technologies within generative AI economics \citep{lazaroiu2024digital}.

The introduction of Bidirectional Encoder Representations from Transformers (BERT) by \citet{devlin2018bert} marked a significant advance in the field of NLP. BERT leverages transformer architectures to generate contextualized word embeddings, improving the understanding of linguistic nuances. Using these contextualized word embeddings, BERT has shown large improvements across various NLP tasks compared to more traditional NLP approaches and models \citep[e.g.,][]{borchert2024industry, bressem2024medbert, catelli2022deceptive}. BERT’s success can be attributed to its bidirectional training of transformers, which enables it to consider both the left and right context of a word simultaneously, leading to a deeper understanding of the nuances in text. Unlike traditional models that rely on fixed word embeddings or simpler context windows, BERT dynamically adjusts its word representations based on the surrounding words in a sentence.

Recent studies highlight the growing importance of NLP in extracting valuable insights from textual data across multiple business functions, such as analyzing online consumer reviews \citep{catelli2022deceptive}, understanding medical documents \citep{bressem2024medbert}, taxonomic classification of DNA sequences \citep{helaly2022bert}, or capturing domain knowledge \citep{chen2024supplementing}. These applications demonstrate the potential of NLP in streamlining processes and improving decision-making in various domains.

Drawing inspiration from the success of BERT models in various domains, we introduce CareerBERT, a fine-tuned language model designed for job recommendations. CareerBERT enables the representation of both resumes and recommendable items (jobs) in a shared embedding space, facilitating the identification of the most suitable job for a given resume from our database. 

To create a comprehensive representation of jobs, we combine data from the European Skills, Competences, Qualifications and Occupations (ESCO)\footnote{\href{https://esco.ec.europa.eu/en}{https://esco.ec.europa.eu/en}} taxonomy and job advertisements from the EURopean Employment Services (EURES)\footnote{\href{https://eures.europa.eu/index_en}{https://eures.europa.eu/index\_en}} online job platform. The ESCO taxonomy provides a structured framework for describing jobs, while the EURES job advertisements offer real-world examples of job descriptions. By leveraging both sources, we aim to capture the essential characteristics of jobs and enable accurate matching with resumes. That is, while previous approaches primarily focus on job recommendations based on a fixed set of concrete job advertisements, CareerBERT ensures an up-to-date and well-defined representation of general job titles in the labor market due to the focus on standardized ESCO job classifications.

Our recommendation system consists of two key components: a domain-adapted language model and a resume-job matching module. We first investigate the effectiveness of domain adaptation by comparing the performance of different base models, such as GBERT and jobGBERT, with and without domain-adaptive pre-training. The best-performing model is then fine-tuned on the downstream task of matching resumes with jobs, with ESCO and EURES data serving as the training data.

For the resume-job matching module, we employ a Siamese network architecture with a Multiple Negatives Ranking (MNR) loss function \citep{reimers2019sentence}. The MNR loss function works with sentence pairs, specifically job titles paired with their corresponding skills, descriptions, or synonyms, and calculates cosine similarities between their embeddings. This loss function is specifically designed to minimize the distance between positive pairs while implicitly pushing away negative examples through its ranking mechanism. This approach allows us to learn a shared embedding space where semantically similar resumes and jobs are placed close to each other. During training, the model learns to minimize the distance between matching resume-job pairs while maximizing the distance between non-matching pairs. This enables the model to effectively capture the semantic similarities between resumes and jobs, facilitating accurate job recommendations.

We follow a two-step evaluation approach to comprehensively assess CareerBERT's performance, consisting of an application-grounded evaluation using job advertisements and a human-grounded evaluation using real-world resumes and Human Resources (HR) expert feedback \citep{doshi2017towards}. The results confirm CareerBERT's effectiveness in supporting career consultants by generating relevant job recommendations based on resumes.

This paper makes several notable contributions to the field of NLP and job recommendation systems. First, we introduce a comprehensive and innovative approach to corpus creation that combines data from the ESCO taxonomy and EURES job advertisements, allowing for a more accurate and up-to-date representation of the labor market. Second, we demonstrate the effectiveness of domain adaptation in the context of job recommendation systems, with CareerBERT outperforming both traditional and state-of-the-art embedding approaches in systematic evaluations. Third, our two-step evaluation approach, consisting of an application-grounded evaluation using job advertisements and a human-grounded evaluation using real-world resumes and expert feedback, provides a comprehensive assessment of CareerBERT's performance and demonstrates its robust effectiveness across diverse professional backgrounds. This evaluation methodology can inform future research and development of similar recommendation systems. Finally, we demonstrate the effectiveness of several integrated preprocessing techniques, such as internal relevance classifiers to reduce noise and improve the overall performance.

The rest of the paper is structured as follows: Section \ref{sec:related_work} reviews the relevant literature on AI and NLP applications in the job market and job consultation fields. Section \ref{sec:methods} describes our proposed methodology of CareerBERT, including its data preparation, model pre-training, and fine-tuning processes. Section \ref{sec:results} presents our evaluation results by assessing CareerBERT's performance and its reception by HR experts. Finally, Section \ref{sec:discussion} discusses findings and implications of our work for research and practice, acknowledges the limitations of the current study, and outlines avenues for future research.

\section{Foundations and Related Work}
\label{sec:related_work}

The rapid evolution of the modern job market, driven by technological advancements and the growing demand for effective job matching, has led to significant research in the fields of NLP and recommender systems. This section provides an overview of the key concepts, methodologies, and findings from relevant studies that inform our proposed approach, CareerBERT, a fine-tuned BERT model designed to support job seekers and career counselors in the job-matching process.

\subsection{NLP Techniques for Job Matching}
\label{subsec:nlp_job_matching}

The emergence of advanced NLP techniques has revolutionized the field of job matching, enabling more accurate and efficient pairing of job seekers with relevant opportunities. Traditional methods like Term Frequency-Inverse Document Frequency (TF-IDF) were widely used for text representation in various NLP tasks, including job matching \citep{zhang2010using}. TF-IDF assigns weights to words based on their frequency within a document and their rarity across a corpus, allowing for the identification of important terms \citep{ramos2003using}. While TF-IDF has been effective in capturing lexical similarities, it lacks the ability to capture semantic relationships between words and documents, which is crucial for understanding the context and meaning in job-related data. 

Embeddings have emerged as a fundamental concept in NLP, playing a crucial role in capturing semantic relationships between words, sentences, and documents \citep{siebers2022survey, catelli2022deceptive}. In the context of job matching and recommender systems, embeddings allow job-related data, such as resumes, job advertisements, and jobs, to be represented in a dense, low-dimensional vector space called the embedding space \citep{mikolov2013efficient, pennington2014glove}. This representation facilitates the computation of similarity measures between job seekers and job opportunities, forming the basis for effective job matching algorithms \citep{zhao2021embedding, bhatia2019end}.

Recent advancements in unsupervised sentence embedding learning have led to the development of more sophisticated techniques, such as the Transfor-mer-based Sequential Denoising Auto-Encoder (TSDAE) proposed by \citet{wang2021tsdae}. TSDAE leverages the power of Transformer-based language models to learn meaningful sentence representations by reconstructing original sentences from corrupted input sequences. This approach enables the model to capture rich semantic information and generate high-quality sentence embeddings without the need for labeled data, suggesting its potential for improving job matching and recommendation systems.

BERT \citep{devlin2018bert}, a pre-trained language model, has achieved state-of-the-art performance on various NLP tasks \citep[e.g.,][]{catelli2022deceptive, helaly2022bert, bressem2024medbert}. This success can be attributed to BERT’s use of a deep bidirectional transformer architecture, which allows it to learn contextualized word representations by considering both left and right contexts during training. Additionally, BERT’s pre-training on large text corpora using masked language modeling and next sentence prediction tasks enables it to capture complex linguistic patterns and semantic relationships, which can be fine-tuned for a wide range of downstream applications, such as text classification, named entity recognition, and machine translation. However, BERT is less suitable for directly comparing the semantic similarity between entire sentences or documents. To address this limitation, Sentence-BERT (SBERT) \citep{reimers2019sentence} adapts the BERT architecture to generate semantically meaningful sentence embeddings using a Siamese network architecture with a bi-encoder structure. The Siamese network consists of two identical BERT models that share the same weights and are trained to encode two input sentences independently. The training objective of SBERT is to maximize the cosine similarity between embeddings of semantically similar sentences while minimizing the similarity between dissimilar sentences. This architecture allows SBERT to efficiently compute sentence embeddings that capture semantic similarity. SBERT can utilize any BERT model as its base model, including domain-specific or language-specific BERT variants, making it particularly well-suited for efficiently assessing semantic similarity in fixed corpus scenarios, such as matching resumes to jobs.

Recent studies have explored various applications of NLP techniques in the job domain, leveraging job-related data sources such as job codes and descriptive information from the ESCO taxonomy. For example, \citet{li2023skillgpt} developed a system for extracting skills from job descriptions by converting ESCO entries into vectorized representations, enabling a vector-based similarity search between semantically related concepts.

\subsection{Adapting Language Models for the Job Domain}
\label{subsec:domain_specific_lm}

While BERT is not a domain-specific model, it can be pretrained on domain-specific data to capture the nuances and semantics of a particular domain. In the job market context, domain-specific language models have the potential to significantly improve the accuracy and relevance of job matching and recommendation systems.

Several researchers have focused on developing domain-specific language models tailored to the HR and recruiting space. For example, jobBERT uses a BERT model pre-trained on job advertisements to obtain vector representations of job titles \citep{decorte2021jobbert}. Similarly, \citet{gnehm2022evaluation} developed jobGBERT, a German-language version trained on millions of job ads. These models showcase the benefits of transfer learning in building NLP tools that capture the unique language and semantics of the job domain.

CareerBERT's training follows the framework proposed by \citet{gururangan2020don}, which distinguishes between different pre-training stages:

\begin{flushleft}
    \hspace{1.47cm} pre-training (general domain) \\
    \hspace{1cm} + domain-adaptive pre-training (downstream domain) \\
    \hspace{1cm} + task-adaptive pre-training (downstream task). \\
\end{flushleft}

Domain-adaptive pre-training has emerged as a popular approach to improve the performance of language models on specific downstream tasks. It involves pre-training a model multiple times, first on a general domain and then on a domain specific to the downstream task. This approach has been studied for various tasks and has shown promising results, such as for instance in adapting a general model to the domain of climate reporting \citep{webersinke2021climatebert}.

For our proposed CareerBERT model, we compare the effectiveness of jobGBERT, a domain-specific BERT model pre-trained on German job advertisements, and GBERT \citep{chan2020gbert}, a general German BERT model, in our job matching task. These models serve as the foundation for our CareerBERT model, which we discuss in detail in Section \ref{sec:model_selection}.

\subsection{Advances in NLP-Driven Job Matching}
\label{subsec:advances_nlp_job_matching}

A variety of studies have focused on directly matching job seekers with relevant opportunities using NLP techniques. Table~\ref{tab:relatedwork} provides an overview of relevant work in this field. For example, \citet{yang_combining_2017} proposed a hybrid job recommendation system based on statistical relational learning. For matching jobs and users, they extracted a variety of attributes from user resumes and job postings, such as job classes, user skills, and the history of most recent companies. \citet{dave_combined_2018} developed a representation learning framework that utilizes the information of different historical job data parsed from textual resume data to jointly learn the representation of jobs and skills in a shared embedding space. On this basis, it was possible to offer improved job and skill recommendations based on available job ads. Instead of extracting attributes from job-related documents, \citet{elsafty_document-based_2018} used a Doc2Vec approach to represent textual job advertisements as dense vectors. To recommend job postings to users in an online job market platform, the authors considered previous user interactions with other job postings based on user bookmarks and reply intentions.

More recently, \citet{zhao2021embedding} developed a deep learning model for CareerBuilder that treated resumes as a query to retrieve and rank relevant job listings, identifying skills and job titles as the most important features for modeling similarity. Similarly, \citet{bhatia2019end} investigated matching candidate profiles to jobs under the assumption that resumes and job descriptions should contain similar task information, using a BERT-based model to predict match scores between candidates and job offers. Another notable advancement is ESCOXLM-R by \citet{zhang2023escoxlm}, which introduces a novel approach using XLM-RLarge as a language model backbone with domain-adaptive pre-training on the ESCO taxonomy to enable cross-lingual job and skill classification. The work that is most comparable to ours is that from \citet{lavi2021consultantbert}. They proposed conSultantBERT, a fine-tuned SBERT model, which aims to construct useful embeddings of textual information in resumes and job ads to exploit a shared dense feature representation for matching jobs and job seekers. However, conSultantBERT, has been trained on recruiter-defined resume-job pairs, which carries the risk of introducing human bias into the recommendations.

Our approach also builds on the general idea of exploiting the similarity between resumes and job characteristics in a shared embedding space. However, instead of matching resumes to a fixed set of concrete job advertisements, which constitutes the general foundation of all existing proposals summarized in Table~\ref{tab:relatedwork}, we focus on matching resumes to general job categories derived from the well-defined ESCO taxonomy. This approach aims to mitigate human biases that may be present in historical job advertisements as described by \citet{hardy2022bias}.

Hence, CareerBERT extends the ideas from related studies while introducing novel approaches for taxonomy-based matching and bias reduction. By leveraging the ESCO taxonomy and focusing on matching resumes to job categories rather than specific historical postings, our approach addresses the limitations of existing job matching systems, such as the reliance on outdated frameworks and the failure to fully utilize the rich data embedded in resumes and job listings \citep{musset2018working}. Additionally, we incorporate data from EURES job advertisements to ensure that our model is always up-to-date, as EURES job advertisements reflect the current job opportunities and complement the ESCO taxonomy, which is only periodically revised.

In summary, our proposed CareerBERT model builds upon the foundations of NLP, leverages recent advances in domain-specific language models, and introduces novel approaches to address the limitations of existing job matching systems. By incorporating insights from relevant studies, adopting the training framework proposed by \citet{gururangan2020don}, and focusing on taxonomy-based matching and bias reduction, CareerBERT has the potential to innovate the job matching process and contribute significantly to the field of NLP-driven job recommender systems.

\section{Development of Job Recommendation System with CareerBERT}
\label{sec:methods}

In this section, we describe our proposed job recommendation system based on CareerBERT by outlining its core functionality and constituent components.

\subsection{Overview of the Job Recommendation System based on CareerBERT}

CareerBERT lies at the heart of our embedding-based recommendation system specifically designed to facilitate job search and recommendation. It transforms both resumes and job descriptions into high-dimensional embeddings that capture semantic meaning. These embeddings are represented as vectors in a shared space, enabling job matching through cosine similarity computations.

\begin{landscape}
\scriptsize
\begin{longtable}{@{}>{\raggedright\arraybackslash}p{3cm}>{\raggedright\arraybackslash}p{3cm}>{\raggedright\arraybackslash}p{3cm}>{\raggedright\arraybackslash}p{5cm}>{\raggedright\arraybackslash}p{5cm}@{}}
\caption{Comparison of different approaches in job recommendation and skill extraction.
\label{tab:relatedwork}} \\

\toprule
\textbf{Approach} & \textbf{Input} & \textbf{Architectural Backbone} & \textbf{Output} & \textbf{Evaluation Approach} \\
\midrule
\endfirsthead

\multicolumn{5}{c}%
{{\tablename\ \thetable{}} -- \textit{continued from previous page}} \\
\toprule
\textbf{Approach} & \textbf{Input} & \textbf{Architectural Backbone} & \textbf{Output} & \textbf{Evaluation Approach} \\
\midrule
\endhead

\midrule
\multicolumn{5}{r}{\textit{Continued on next page}} \\
\endfoot

\bottomrule
\endlastfoot

\citet{yang_combining_2017} & Structured attributes from user resumes and job advertisements & Hybrid recommendation system based on cost-sensitive statistical relational learning & Job recommendations based on available job ads for resume inputs; candidate recommendations based on available resumes for job ads & Comparison of different backbone models and configurations \\[2ex]

\citet{dave_combined_2018} & Parsed attributes from textual user resumes & Joint Margin network embedding framework & Job and skill recommendations based on available job descriptions & Application in real-world case studies and comparative evaluation with baseline models \\[2ex]

\citet{elsafty_document-based_2018} & Textual job advertisements and past user interactions with job ads & Word2Vec / Doc2Vec & Job recommendations based on available job ads & Comparison of different backbone models and configurations and application in real-world case study \\[2ex]

\citet{bhatia2019end} & Textual job descriptions/ advertisements & BERT & Candidate recommendations based on available resumes & Simple model evaluation \\[2ex]

CareerBuilder \citep{zhao2021embedding} & Textual user resumes or job advertisements & Deep learning embedding model with convolutional layer and attention layer & Job recommendations based on available job ads for resume inputs; candidate recommendations based on available resumes for job ads & Application in real-world case study with human-grounded evaluation \\[2ex]

jobBERT \citep{decorte2021jobbert} & Titles of job advertisements + associated skills & BERT & Job title normalization (backbone functionality) & Comparison of different backbone models and configurations \\[2ex]

jobGBERT \citep{gnehm2022evaluation} & Titles of job advertisements + associated skills & jobBERT & Job title normalization (backbone functionality) & Comparison of different backbone models and configurations \\[2ex]

conSultantBERT \citep{lavi2021consultantbert} & Textual user resumes or job advertisements & Sentence-BERT & Job recommendations based on available job ads for resume inputs; candidate recommendations based on available resumes for job ads & Comparison of different backbone models and configurations \\[2ex]

SkillGPT \citep{li2023skillgpt} & Textual job advertisements/ user resumes & Vicuna-13B & Skill extraction based on standardized ESCO taxonomy & - \\[2ex]

ESCOXLM-R \citep{zhang2023escoxlm} & ESCO taxonomy data covering 27 languages & XLM-RLarge with domain-adaptive pre-training & Multilingual job and skill classification based on ESCO taxonomy & Comprehensive evaluation across 9 job-related datasets in 4 languages \\[2ex]

LLM-based GANs Interactive Recommendation \citep{du_enhancing_2024} & Textual user resumes & ChatGLM-6B + SIM-BERT + GANs & Job recommendations based on available job ads & Comparative evaluation with a variety of baseline models \\[2ex]

\textbf{CareerBERT (our proposal)} & \textbf{Textual user resumes} & \textbf{Sentence-jobGBERT} & \textbf{Career recommendations based on standardized ESCO job classifications} & \textbf{Comparison of different backbone models and configurations and
human-grounded evaluation} \\
\end{longtable}
\end{landscape}

\begin{figure}[h!]
    \centering    \includegraphics[width=0.9\linewidth]{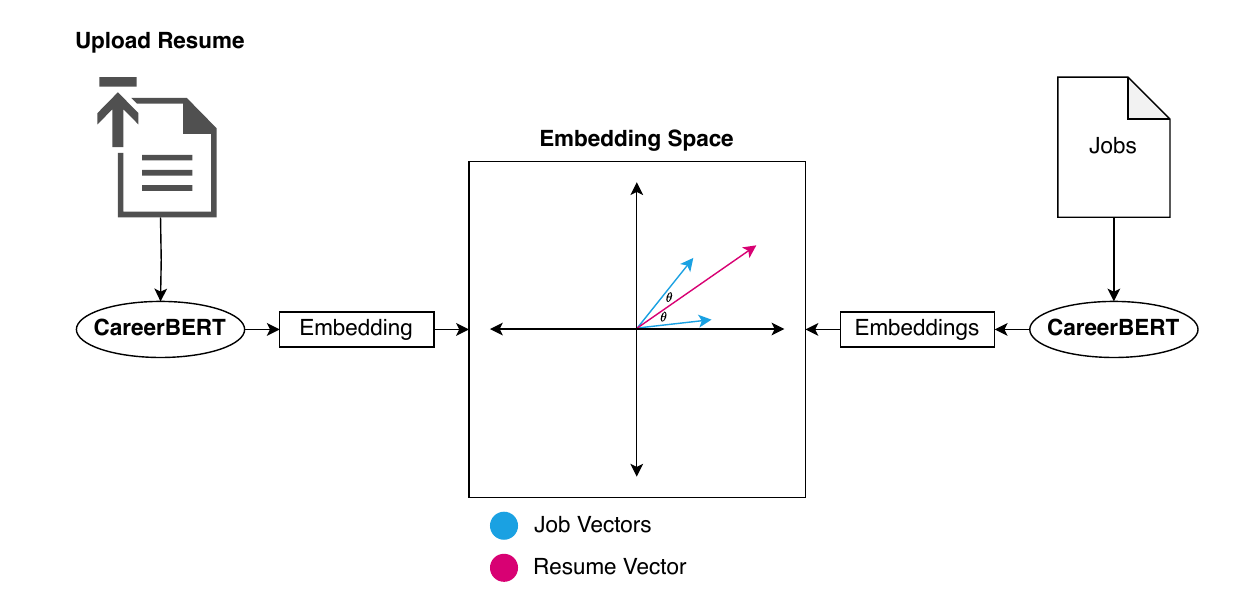}
    \caption{Workflow of the proposed job recommendation system with CareerBERT at its core. A user's resume is transformed into a high-dimensional embedding (Resume Vector) through CareerBERT's embedding process. Similarly, jobs are converted into embeddings (Job Vectors) in the same space. The system then finds suitable job recommendations by measuring the cosine similarity between the vectors, where smaller angles ($\theta$) indicate higher similarity.}
    \label{fig:workflow}
\end{figure}

The workflow of our job recommendation system is visualized in Figure~\ref{fig:workflow} and can be summarized in the following steps:
\begin{enumerate}
    \item \textbf{User input}: The user uploads their resume through the application interface (see \ref{app:interface} for details), which serves as the initial query for the recommender system.
    \item \textbf{Resume embedding}: CareerBERT processes the user's resume and generates a high-dimensional embedding that captures the semantic content of the resume.
    \item \textbf{Job embedding}: Similarly, CareerBERT creates embeddings for each job in its corpus. These embeddings encapsulate the essential characteristics and requirements of each job.
    \item \textbf{Similarity computation}: The system computes cosine similarity between the resume and job embeddings, represented as vectors. Cosine similarity measures the angle between these vectors, providing scores ranging from -1 to 1, where 1 indicates perfect semantic alignment (vectors pointing in the same direction), 0 indicates no semantic relationship (perpendicular vectors), and -1 suggests semantic opposition (vectors pointing in opposite directions).
    \item \textbf{Recommendation generation}: Based on the cosine similarity scores, CareerBERT identifies the jobs that are most semantically similar to the user's resume. These jobs are then presented to the user as personalized recommendations through the application interface~(\ref{app:interface}).
\end{enumerate}

By representing both resumes and jobs as embeddings in the same high-dimensional space, our proposed job recommendation system can uncover meaningful connections that might be missed by traditional keyword matching. This enables the system to recommend jobs that align with the user's skills, experiences, and interests, even if the job titles or descriptions use different terminology.

It is important to note that the effectiveness of our embedding-based recommender system depends on the quality of the embeddings generated by CareerBERT. No single model can achieve perfect performance across all use cases, and fine-tuning the model on domain- and task-specific data, such as resumes and job descriptions, is often necessary to optimize its performance for a particular application.

\subsection{Preprocessing of Job Data}
\label{sec:preprocessing_job_data}

In order to improve the performance of our job recommendation system, we use various state-of-the-art approaches to preprocess the corpus of job-related data. Initially, the corpus consisted solely of job descriptions from the ESCO database. While ESCO provides a standardized and well-structured representation of the European labor market, relying exclusively on this taxonomy data presents several limitations, such as:

\begin{itemize}
    \item ESCO data is only revised periodically, which means that it may not always reflect the most up-to-date job market trends and requirements.
    \item Taxonomies, by nature, are simplified representations of reality and may not capture the full complexity and nuances of real-world job data.
    \item Full reliance on ESCO data limits CareerBERT's ability to adapt quickly to changes in the job market, such as the emergence of new job titles, skills, or industry-specific terminology.
\end{itemize}

\subsubsection{Incorporating EURES Job Advertisements}
\label{sec:eures_job_ads}

To address the limitations of relying solely on ESCO data, we identified the EURES database as a promising addition. The EURES database contains a large collection of real job advertisements from across Europe, with the job advertisements annotated with corresponding ESCO job titles, providing a natural link between the two data sources.

By incorporating EURES job advertisements into the corpus, we aim to capture the most recent trends and requirements in the job market, as the EURES database is continuously updated with new job advertisements. Additionally, this provided a more diverse and representative set of job characteristics, reflecting the actual language and terminology used by employers in their job advertisements. Finally, reducing the system's dependence on the ESCO taxonomy allows CareerBERT to adapt more quickly to changes in the job market.

While the EURES job advertisements provide valuable real-world data for training CareerBERT, they also introduce several challenges, including wrongly labeled job ads, noisy content (e.g., recruiter/company names, benefits), and the maximum sequence length limitation of BERT (512 tokens). To address these issues, we developed a classifier for job ad shortening, which is detailed in \ref{app:classifier}.

We developed a binary text classification model designed to identify and extract relevant information from job advertisements, focusing specifically on key sections such as job titles, job descriptions, and job requirements. To create a labeled dataset, we manually annotated 400 job ads, distinguishing between relevant paragraphs (containing details about job roles and qualifications) and non-relevant paragraphs (containing company information, legal disclaimers, or promotional content). For the classification task, we utilized the jobGBERT base model from \citep{gnehm2022evaluation} and fine-tuned it on our annotated dataset. The model was trained with a binary cross-entropy loss function.

The benefits of employing this classifier are threefold. First, it reduces the text length to fit within BERT's sequence length limitations. Second, it focuses on job-specific information, minimizing noise and irrelevant content. Third, it helps to remove outliers and wrongly labeled jobs, improving the overall data quality.

\subsubsection{Job Advertisement Centroids and Job Centroids}
\label{sec:jacs_and_jcs}

Our goal is to create a comprehensive representation of the 3,008 unique jobs defined in the ESCO taxonomy. To achieve this, we leverage two main data sources: job advertisements from the EURES platform and the ESCO taxonomy itself.

First, we generate embeddings for each relevant part (job title, job description, job requirements) of an EURES job advertisement using the fine-tuned CareerBERT models. Each EURES job advertisement comes with a single annotated ESCO job ID, which provides a natural binary relevance scheme for training and evaluation - the annotated ESCO job is considered the single relevant match for that advertisement, while all other ESCO jobs are treated as non-relevant. Then, we calculate job advertisement centroids by averaging the job advertisement embeddings for each ESCO job, following the approach proposed by \citet{schopf2021lbl2vec}. These job advertisement centroids provide a valuable representation of real-world job data.

However, relying solely on job advertisement centroids has some limitations. The job advertisement texts may contain noise, leading to data quality issues. Additionally, EURES job advertisements were not available for all 3,008 ESCO jobs, resulting in incomplete coverage of our corpus.

To address these limitations and leverage the strengths of both data sources, we develop a hybrid approach by calculating job centroids. Job centroids are created by combining the job advertisement centroids with the embeddings of the best-performing ESCO information (ESCO job descriptions). This approach allows us to incorporate the structured and reliable ESCO data while benefiting from the real-world job market representation provided by the job advertisement centroids.

By combining these two data sources, we mitigate the risk of misclassified job recommendations and leverage the complementary nature of the ESCO taxonomy and EURES job advertisements. The resulting job centroids serve as the final embeddings that form the embedding space used by CareerBERT for job recommendations.

This comprehensive approach to corpus creation distinguishes our research from others in the field. By carefully curating and processing data from both the ESCO taxonomy and EURES job advertisements, and by employing advanced techniques such as classifier training, we create a robust, relevant, and representative corpus that forms the foundation for CareerBERT's job recommendation capabilities.

In summary, our goal is to create embeddings for the 3,008 unique ESCO jobs. We achieve this by calculating job advertisement centroids from EURES data and combining them with the best-performing ESCO information to create job centroids. These job centroids form the embedding space used by CareerBERT for job recommendations, leveraging the strengths of both real-world job advertisements and the structured ESCO taxonomy.

\section{Results}
\label{sec:results}

The results of our experiments and evaluations provide valuable insights into the performance and effectiveness of CareerBERT as an NLP-based support tool for career consultants. In this section, we present the experimental setup, findings from our application-grounded evaluation, which assessed the model's ability to recommend relevant ESCO jobs based on EURES job advertisements, as well as the results from our human-grounded evaluation, which involved real-world resumes and expert feedback from HR professionals.

\subsection{Experimental Setup}
\label{sec:model_selection}

Selecting an appropriate model architecture is crucial for CareerBERT's success in creating high-quality embeddings from both user queries (resumes) and corpus items (jobs). We chose SBERT \citep{reimers2019sentence}, an adaptation of BERT using a bi-encoder structure, as it is particularly well-suited for efficiently assessing semantic similarity in fixed corpus scenarios, such as matching resumes to jobs.

\subsubsection{Base Model Selection}
\label{sec:base_model_selection}

SBERT relies on BERT base models for its internal representation. To ensure compatibility with the German language and the job domain, we considered two BERT variants as base models:

\begin{enumerate}
    \item \textbf{GBERT} \citep{chan2020gbert}: A German BERT model pre-trained on a large corpus of German text data, serving as a strong baseline for general German language understanding. GBERT provides a solid foundation for capturing the semantics of the German language, which is essential for processing resumes and job descriptions written in German.
    \item \textbf{jobGBERT} \citep{gnehm2022evaluation}: An extension of GBERT that has undergone additional domain-adaptive pre-training on four million German-speaking job ads from Switzerland (1990-2020). By incorporating domain-specific knowledge from the job market, jobGBERT has the potential to better understand the nuances and terminology used in resumes and job descriptions, making it potentially more suitable for the job recommendation task.
\end{enumerate}

The choice between GBERT and jobGBERT allows us to investigate the impact of domain-adaptive pre-training on CareerBERT's performance in generating relevant job recommendations. By comparing these two models, we can assess whether the additional domain-specific knowledge acquired by jobGBERT translates into improved job recommendations compared to the general language understanding provided by GBERT.

Furthermore, evaluating both base models within the SBERT architecture enables us to determine the optimal configuration for CareerBERT. The combination of the most suitable base model and the SBERT architecture will help ensure that CareerBERT can effectively capture the semantic similarity between resumes and jobs, ultimately leading to fast and relevant job recommendations for users.

In the following sections, we will discuss the training data selection and the training process employed to fine-tune the selected base models within the SBERT architecture, further enhancing their ability to generate meaningful embeddings for resumes and jobs.

\subsubsection{Training Data Selection}
\label{sec:train_data}

To ensure the quality and consistency of our training data, we relied exclusively on the ESCO database. ESCO provides a well-curated and standardized representation of the European labor market, making it an ideal source for training CareerBERT while minimizing the risk of incorporating biases that may be present in real-world job application data \citep{hardy2022bias}.

As ESCO job data is not readily available in the required format for training SBERT models, we created sentence pairs by combining job titles with their associated information, such as skills, synonyms, and descriptions, as shown in Table~\ref{tab:sentence_pairs}. This approach has proven effective in similar contexts \citep{lavi2021consultantbert}.

\begin{table}[h!]
\centering
\footnotesize
\begin{tabular}{lr}
\toprule
Sentence Pair & Count \\
\midrule
job title – skill & 113,685 \\
job title – synonym & 14,723 \\
job title – description & 2,937 \\
\bottomrule
\end{tabular}
\caption{Number of sentence pairs.}
\label{tab:sentence_pairs}
\end{table}

\subsubsection{Training Process}
\label{sec:training_process}

\begin{figure}[h!]
\centering
\includegraphics[width=1\linewidth]{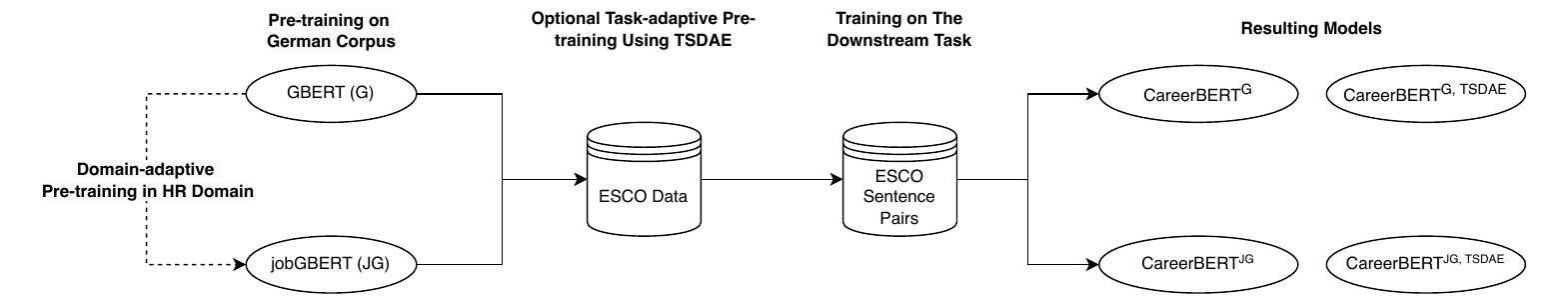}
\caption{Overview of CareerBERT's training process. Starting with either GBERT or jobGBERT (a domain-adapted model for Human Resources (HR) \citep{gnehm2022evaluation}) as base models, we explored variants with and without task-adaptive pre-training using Transformer-based Denoising AutoEncoder (TSDAE) before final training on ESCO sentence pairs.}
\label{fig:training_process}
\end{figure}

Our pre-training process for CareerBERT follows the framework proposed by \citet{gururangan2020don}, which consists of several stages: pre-training on general domain data, domain-adaptive pre-training in the HR domain, and task-adaptive pre-training. Figure \ref{fig:training_process} provides an overview of our training process.

To investigate the impact of different pre-training strategies, we evaluated two base models: GBERT and jobGBERT (the domain-adaptive pretrained model). Furthermore, we assessed the effect of applying task-adaptive pre-training using a Transformer-based Sequential Denoising Auto-Encoder (TSDAE) approach \citep{wang2021tsdae}. TSDAE is an unsupervised learning method that involves masking parts of the input data and training the model to reconstruct the missing segments, enabling it to capture text semantics without requiring sentence pairs.

Following the pre-training stages, we fine-tuned the models on the downstream task of matching resumes and jobs using sentence pairs created from the ESCO data. These sentence pairs were generated by combining job titles with their associated information, such as skills, synonyms, and descriptions, as shown in Table~\ref{tab:sentence_pairs}. Although ESCO differentiates between essential and optional skills, we treated both as equally important and concatenated them for training purposes.

Our training approach resulted in four different models:

\begin{enumerate}
    \item[(1a)] CareerBERT\textsuperscript{G}: GBERT base without TSDAE\footnote{Available here: \href{https://huggingface.co/lwolfrum2/careerbert-g}{https://huggingface.co/lwolfrum2/careerbert-g}}
    \item[(1b)] CareerBERT\textsuperscript{G, TSDAE}: GBERT base with TSDAE
    \item[(2a)] CareerBERT\textsuperscript{JG}: jobGBERT base without TSDAE\footnote{Available here: \href{https://huggingface.co/lwolfrum2/careerbert-jg}{https://huggingface.co/lwolfrum2/careerbert-jg}}
    \item[(2b)] CareerBERT\textsuperscript{JG, TSDAE}: jobGBERT base with TSDAE
\end{enumerate}

By evaluating these models, we aimed to assess the impact of domain-adaptive pre-training (jobGBERT vs. GBERT) and the effectiveness of TSDAE on the performance of CareerBERT in generating relevant job recommendations.

\subsection{Evaluations and Metrics}
\label{sec:evaluation_approach}

Evaluating the performance of CareerBERT is crucial for understanding its effectiveness and identifying areas for improvement. We conducted both an application-grounded evaluation to identify the best-performing model configuration and a human-grounded evaluation to assess the effectiveness of CareerBERT's recommendation system.

\subsubsection{Application-grounded Evaluation}
\label{sec:application_grounded_eval}

After fine-tuning our BERT base models (i.e., GBERT and jobGBERT, as described in Section \ref{sec:base_model_selection}) to capture the semantics of resumes, we need to evaluate their performance and determine which model is best suited for generating meaningful embeddings of resumes and job characteristics. Ideally, we would use a test set of resumes annotated with corresponding ESCO IDs. However, due to the lack of such a dataset, we used an alternative approach.

We created a test set with 2,250 job advertisements from EURES, which are classified by ESCO IDs. These job advertisements serve as a suitable proxy for evaluating the models' ability to match resumes to jobs. For the test corpus, we used the ESCO job descriptions, as they provide a standardized and comprehensive representation of the European job market.

To evaluate the trained models, we first created an embedding space using the ESCO job descriptions as the corpus. We then generated embeddings for the test job advertisements from EURES. Using SBERT's RerankingEvaluation, we implemented a reranking approach that considers the full spectrum of cosine similarity scores (-1 to 1) between these embeddings. This approach computes Mean Reciprocal Rank at 100 (MRR@100) across the complete similarity range to measure the models' performance in matching the test job advertisements to the correct ESCO job descriptions.

The MRR@100 assesses the models' ability to rank the correct item (in this case, the matching ESCO job) highly within the top 100 results. A higher MRR@100 score indicates better performance. For example, if the correct ESCO job is ranked at position 5 in the recommendation list for a given job advertisement query, the reciprocal rank would be $\frac{1}{5} = 0.2$. If the correct ESCO job code is ranked at position 20 for another query, the reciprocal rank would be $\frac{1}{20} = 0.05$. The MRR@100 is then calculated as the average of these reciprocal ranks across all queries.

By using MRR@100, we can effectively evaluate the model's ability to retrieve the correct ESCO job code for a given job advertisement and assess the effectiveness of our training approach. This evaluation method allows us to compare the performance of different models and select the one that best matches job advertisements to ESCO job descriptions.

\textbf{Experiment 1: Job Advertisement Truncation.} Experiment 1 aimed to test the assumption that truncating job advertisements using a classifier improves embedding quality. This idea is based on the limitation of BERT-based models to have a limited number of tokens that can be processed. We trained a classifier on annotated job advertisements to extract relevant parts (job title, job description, and job requirements). The rationale behind this approach lies in reducing noise and focusing on the most informative segments of the text, which aligns with the principles of efficient feature selection in NLP tasks. By prioritizing semantically rich content, we aimed to enhance the model’s ability to capture meaningful patterns during embedding generation.

The results in Table \ref{tab:experiment1} show that truncation of job advertisements using a classifier achieves considerably higher MRR@100 scores than the method of cutting off job advertisements at the token limit. This highlights the importance of an informed truncation strategy in NLP pipelines, as it helps preserve critical information while avoiding the dilution of embeddings with irrelevant content.

\begin{table}[ht]
     \centering
     \footnotesize
     \begin{tabular}{lcc}
         \toprule
         & \multicolumn{2}{c}{Truncation Method} \\ \cmidrule{2-3}
         Model  & cut-off at token limit & classifier  \\ \midrule
         CareerBERT\textsuperscript{JG} & 0.307  & \textbf{0.328}  \\
         CareerBERT\textsuperscript{G} & 0.290  & \textbf{0.312}  \\ \bottomrule
     \end{tabular}
     \caption{Comparison of MRR@100 scores for job advertisements truncated using a classifier versus job advertisements truncated at the token limit, evaluated on CareerBERT with GBERT and jobGBERT base models.}
     \label{tab:experiment1}
\end{table}

\textbf{Experiment 2: Model Comparison.} In this experiment, we compared various approaches for job matching, ranging from traditional embedding methods to state-of-the-art language models. We evaluated expert systems based on Word2Vec and Doc2Vec implementations following \citet{elsafty_document-based_2018}, expert systems conSultantBERT by \citet{lavi2021consultantbert} and ESCOXLM-R by \citet{zhang2023escoxlm}, our CareerBERT models (using GBERT (G) and jobGBERT (JG) as base models), and OpenAI's embedding models (text-embedding-3-small and text-embedding-ada-002\footnote{https://platform.openai.com/docs/guides/embeddings}).

For the experiment, we used ESCO job descriptions as our reference embeddings, since preliminary experiments showed that they consistently provided better matching performance compared to other ESCO data such as skills or job titles. We then evaluated all models by matching classifier-truncated EURES job advertisements to these ESCO job description embeddings. As shown in Table~\ref{tab:experiment2}, CareerBERT with the jobGBERT model achieved the highest MRR@100 score of 0.328, though the margin is narrow compared to OpenAI's models.

\begin{table}[h]
    \centering
    \footnotesize
    \begin{tabular}{lccccc}
        \toprule
        Models & MRR@100 \\ \midrule
        Word2Vec &  0.005 \\
        Doc2Vec (distributed memory) & 0.006 \\
        Doc2Vec (distributed bag of words) & 0.013 \\
        conSultantBERT (reconstructed) & 0.132 \\ 
        OpenAI (text-embedding-ada-002) & 0.302 \\
        CareerBERT\textsuperscript{G} & 0.312  \\
        ESCOXLM-R & 0.312 \\
        OpenAI (text-embedding-3-small) & 0.323 \\ 
        CareerBERT\textsuperscript{JG} & \textbf{0.328} \\ \bottomrule
    \end{tabular}
    \caption{Comparison of the MRR@100 scores for various job matching approaches, including traditional embedding methods, expert systems, and modern language models.}
    \label{tab:experiment2}
\end{table}

Additionally, we conducted a comparative experiment with GPT-4 using a prompt-based matching approach on a random sample of 250 job advertisements. While achieving comparable performance (MRR@100 of 0.324), the chatbot-based approach failed to find valid matches for approximately 33.9\% of cases and required substantially higher computational resources, processing an average of 515,000 input tokens per job advertisement.

The results demonstrate several interesting findings. First, traditional embedding methods (Word2Vec and Doc2Vec variants) perform poorly on this task despite following established approaches from the literature \citep{elsafty_document-based_2018}. Second, conSultantBERT, while using a straightforward BERT-based architecture with mean pooling as proposed by \citet{lavi2021consultantbert}, achieves better results than traditional methods but falls short of modern approaches. Third, both ESCOXLM-R and our CareerBERT models achieve strong performance, with results competitive with or slightly outperforming OpenAI's state-of-the-art embedding models. The success of ESCOXLM-R demonstrates the value of incorporating structured taxonomic knowledge during pre-training, while CareerBERT's performance suggests combining architectural improvements with domain-specific training for specialized tasks.

Furthermore, CareerBERT with the jobGBERT model consistently outperforms CareerBERT with the GBERT model in terms of MRR@100. This performance difference highlights the importance of domain-specific pre-trai-ning, as jobGBERT's initial training on HR-domain text appears to provide advantages even when comparing similarly fine-tuned models. This suggests that a similar pipeline configuration could be beneficial for other domain-specific applications, where models without domain-adaptive pre-training may lack sufficient contextual understanding.

\textbf{Experiment 3: Embedding Comparison.} Experiment 3 evaluated different approaches to represent jobs and their characteristics. First, we truncated the job advertisements using the classifier from Experiment 1. Then, using the best-performing jobGBERT model from Experiment 2, we generated embeddings for both the truncated EURES job advertisements and ESCO features. For EURES data, we created job advertisement centroids by averaging all embeddings that map to the same ESCO occupation code. We compared these centroids against individual ESCO features (job titles, descriptions, and skill sets) to determine which representation captures job-related information most effectively.

While job advertisement centroids demonstrated superior performance, they were only available for 1,700 out of 3,008 ESCO jobs. To balance performance with coverage, we developed a hybrid approach which we called job centroids. These were created by averaging the job advertisement centroids with ESCO job descriptions, which were the best-performing ESCO feature. For occupations without available job advertisement centroids, we used the ESCO job description as the centroid.

\begin{table}[h]
    \centering
    \footnotesize
    \begin{tabular}{
        l
        l
        c
        c
    }
    \toprule
    {Source} & {Embedding Type} & \multicolumn{2}{c}{MRR@100} \\ \cmidrule{3-4}
    & & {Base} & {+TSDAE} \\ \midrule
    EURES & job advertisement centroids & 0.482 & \textbf{0.485} \\[0.1cm] \midrule
    ESCO & job titles & 0.321 & 0.309 \\
    & job descriptions & \textbf{0.328} & 0.298 \\
    & job skill sets & 0.265 & 0.245 \\[0.1cm] \midrule
    EURES \& ESCO & job centroid & \textbf{0.434} & 0.420 \\
    \bottomrule
    \end{tabular}
    \caption{Embedding type comparison using the best-performing jobGBERT model from Experiment 2. Results show MRR@100 scores for different embedding types, comparing the base model with additional Transformer-based Denoising AutoEncoder (TSDAE) pre-training.}
    \label{tab:experiment3}
\end{table}

The results demonstrate that averaged job advertisement centroids consistently outperform individual ESCO features, with the highest MRR@100 score of 0.485 achieved using TSDAE pre-training, compared to scores between 0.245 and 0.328 for ESCO features. While the hybrid job centroids showed a slight decrease in performance (MRR@100 of 0.434 for the base model), this trade-off was accepted as it mitigated the risk of misclassification while maintaining broader coverage across ESCO occupations. The effectiveness of centroids can be attributed to their ability to capture common patterns across multiple job advertisements while reducing the impact of noise in individual postings.

\subsubsection{Human-grounded Evaluation}
\label{sec:human_grounded_eval}

Building upon our experimental findings with job centroid embeddings, we conducted a comprehensive human evaluation of CareerBERT's practical effectiveness. Using CareerBERT with the best-performing jobGBERT model, we performed a human-grounded evaluation using five real-world resumes, as described in \ref{app:resumes}. We generated job recommendations for each resume using CareerBERT\textsuperscript{JG} and then recruited ten HR experts to assess the quality and relevance of these recommendations. The experts were sourced from a range of outplacement and recruitment companies, as well as individuals who work in recruitment roles across different organizations. This diverse panel of experts was carefully selected to provide a comprehensive and informed evaluation of the model's performance in a real-world context. The recruitment process involved leveraging our professional network and utilizing LinkedIn to identify and reach out to suitable candidates.

\begin{figure}[h!]
    \centering
    \includegraphics[width=1\linewidth]{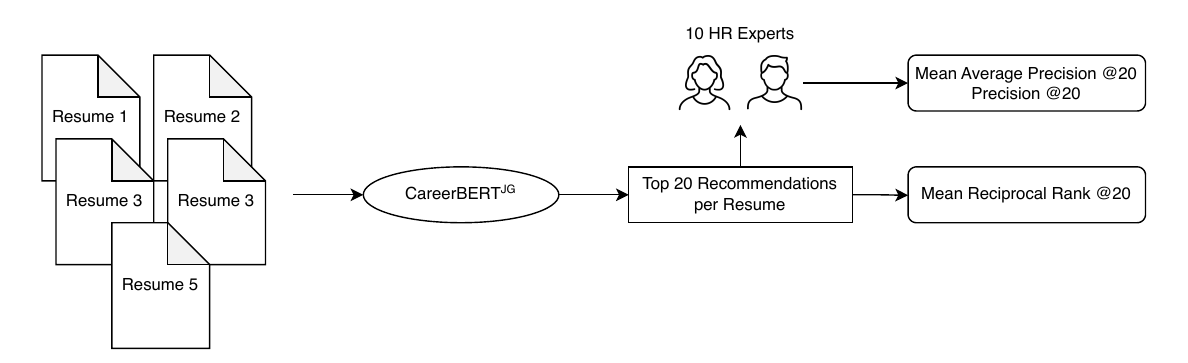}
    \caption{Visualization of human-grounded evaluation. Resumes are processed by CareerBERT\textsuperscript{JG} using the best-performing jobGBERT model and job centroid embeddings}. The model generates the top 20 job recommendations per resume. These recommendations are assessed for relevance by 10 HR experts, who determine whether each recommended job is suitable for the corresponding resume. Based on these expert relevance judgments, we compute three metrics: Mean Average Precision at 20 (MAP@20), Precision at 20 (P@20), and Mean Reciprocal Rank at 20 (MRR@20).
    \label{fig:eval-setup}
\end{figure}

To ensure consistent evaluation across experts, each recommendation was assessed through binary relevance judgments (`relevant' or `irrelevant'). To support informed decisions, experts were provided with standardized ESCO job descriptions for each recommendation. The recommendations were presented in alphabetical order to avoid potential ranking bias. This standardized approach allowed experts to focus on the direct suitability of recommendations based on resume content and job descriptions, rather than incorporating additional factors such as company fit or career transition potential. Figure~\ref{fig:eval-setup} illustrates the evaluation setup, highlighting the involvement of HR experts in assessing the generated job recommendations.

The HR experts evaluated each recommendation's suitability for its corresponding resume, providing valuable feedback on CareerBERT's performance in a real-world setting. While we used in our previous experiments the MRR@100, we limited the human evaluation to the top 20 recommendations per resume and therefore used the MRR@20. This ensured a manageable workload for the experts while maintaining meaningful assessment.

In addition to the MRR@20, we calculated the Mean Average Precision at 20 (MAP@20) and Precision at 20 (P@20) based on the HR experts' relevance assessments. For each recommendation, a job was considered relevant if at least 50\% of the evaluators marked it as suitable for the resume. Using these binary relevance judgments, MAP@20 measures the average precision of the top 20 recommended items across multiple queries or instances, with a MAP@20 close to 1 indicating that the model is more effective at ranking relevant items higher in the recommendation list. Similarly, P@20 measures the proportion of the top 20 recommendations that were deemed relevant by the majority of HR experts, providing a direct measure of recommendation quality. These expert assessments establish the benchmark for relevance, allowing us to assess the real-world applicability and effectiveness of CareerBERT.

To facilitate the human-grounded evaluation process, we developed an interactive web application using Streamlit.\footnote{\href{https://streamlit.io/}{https://streamlit.io/}} The Streamlit application allowed the HR experts to easily view the resumes and their corresponding job recommendations generated by CareerBERT\textsuperscript{JG}. The experts could then rate the suitability of each recommendation directly within the application, providing a user-friendly and efficient way to collect their feedback. The use of the Streamlit application streamlined the evaluation process and ensured a consistent and organized approach to gathering expert assessments.

\textbf{Results of the Human-grounded Evaluation.} The evaluation metrics revealed varying performance across different resumes, as shown in Table~\ref{tab:human_eval_results}.

\begin{table}[htbp]
    \centering
    \footnotesize
    \sisetup{
        table-format=1.3,
        table-alignment=center,
        round-mode=places,
        round-precision=3
    }
    \begin{tabular}{
        l
        S[table-format=1.3]
        S[table-format=1.3]
        S[table-format=1.3]
    }
    \toprule
    {Resume} & {MAP@20} & {P@20} & {MRR@20} \\
    \midrule
    Resume 1 & 0.655 & 0.412 & 1.000 \\
    Resume 2 & 0.310 & 0.160 & 0.303 \\
    Resume 3 & 0.861 & 0.800 & 1.000 \\
    Resume 4 & 0.868 & 0.800 & 1.000 \\
    Resume 5 & 0.862 & 0.625 & 1.000 \\
    \midrule
    Average & 0.711 & 0.559 & 0.861 \\
    \bottomrule
    \end{tabular}
    \caption{Performance metrics of CareerBERT\textsuperscript{JG} using job centroid embeddings across five different resumes. 
    }
    \label{tab:human_eval_results}
\end{table}

CareerBERT\textsuperscript{JG} demonstrated robust performance across most resumes, with particularly strong results for resumes 3, 4, and 5. Notable is the perfect MRR@20 score achieved for four out of five resumes, indicating the model's ability to place highly relevant recommendations at the top of the list. For resume 4, representing a blue-collar professional background, the model achieved the best performance across all metrics, demonstrating its versatility across different occupational domains. Similarly, strong results for resume 5 indicate the model's capability to handle atypical career profiles. For resume 2, being notably shorter, the model achieved significantly lower performance across all metrics, suggesting its sensitivity to input completeness, similar to the cold start problem in traditional recommendation systems \citep{lika2014facing}.

\section{Discussion}
\label{sec:discussion}

The development and evaluation of our job recommendation system, an NLP-based support tool for career counselors that uses CareerBERT to generate meaningful recommendations, has yielded promising results and valuable insights into the application of AI techniques in the field of job recommendation. In this section, we discuss the contributions of our research, its implications for research and practice, as well as its limitations and future research directions.

\subsection{Contribution}
\label{subsec:contribution}

Our research makes several notable contributions to the field of NLP and job recommendation systems. First, we have developed a comprehensive and innovative approach to corpus creation by combining data from the ESCO taxonomy and EURES job advertisements. This approach allows for a more accurate and up-to-date representation of the labor market, as it incorporates both structured taxonomic information and real-world job market data. The utilization of advanced techniques, such as classifier training for job advertisement shortening, further enhances the quality and relevance of the corpus.

Second, our research demonstrates the effectiveness of domain adaptation in the context of job recommendation systems. By comparing the performance of a domain-specific model (jobGBERT) with a general language model (GBERT), and state-of-the-art LLM approaches, we have shown that domain-specific pre-training can lead to improved performance in ranking relevant job recommendations. Our experiments showed that CareerBERT outperformed both OpenAI's embedding models and achieved comparable but more reliable performance than GPT-4 while being significantly more computationally efficient. These findings highlight the importance of considering domain-specific knowledge when developing NLP-based recommendation systems and are in line with research by \citep{gururangan2020don}.

Third, our two-step evaluation approach, consisting of an application-grounded evaluation using job advertisements and a human-grounded evaluation using real-world resumes and expert feedback, provides a comprehensive assessment of CareerBERT's performance following suggestions by \citep{doshi2017towards}. The human-grounded evaluation revealed the model's versatility across different occupational domains, showing strong performance for both traditional profiles and blue-collar backgrounds, while also highlighting important considerations regarding input quality. This evaluation methodology not only assesses our approach's ability to recommend relevant ESCO jobs based on job advertisements but also validates its effectiveness in a practical setting based on the judgment of domain professionals \citep{malik2023employee}.

\subsection{Implications for Research and Practice}
\label{subsec:implications}

The implications of our research extend to both the academic community and practitioners in the field of career consulting and job recommendation.

For researchers, our work demonstrates the potential of leveraging AI techniques, such as BERT and SBERT, for developing effective job recommendation systems. The successful application of domain adaptation and the use of taxonomic data combined with real-world job advertisements provide a foundation for future research in this area. Our findings also underscore the importance of considering task-specific knowledge and the quality of input data when developing recommendation systems, which can guide future research endeavors. Notably, our training data consisted of freely available and reproducible datasets, including ESCO data that exists for multiple languages. This suggests that the success of CareerBERT in the German language could potentially translate to other languages as well, opening up opportunities for further research and application in multilingual settings \citep{chan2020gbert}.

Moreover, our two-step evaluation approach, which incorporates both application-grounded and human-grounded evaluations, offers a comprehensive framework for assessing the performance of recommendation systems. This approach can be adapted and applied to other domains, fostering a more holistic understanding of the effectiveness and real-world applicability of such systems \citep{doshi2017towards}.

Furthermore, our hybrid approach of combining structured ESCO taxonomy data with real-world job advertisements provides distinct advantages for both research and practice. The ESCO taxonomy offers a standardized framework that ensures consistency and reduces ambiguity in job matching, while EURES job advertisements capture current market dynamics and emerging roles. This combination helps mitigate potential biases in historical job advertisements while maintaining practical relevance through real-world data \citep{hardy2022bias}. The successful integration of these complementary data sources demonstrates how structured taxonomic knowledge can enhance job recommendation systems while remaining adaptable to market changes.

Moreover, while large language models show promise for tasks like resume enhancement, our experiments with chatbot-based approaches revealed significant practical limitations. Specifically, when we tested with GPT-4 on a sample of 250 job advertisements, it failed to produce valid matches in 33.9\% of cases, while requiring on average 515,000 input tokens to process a single job advertisement. This computational overhead makes chatbot-based approaches impractical at scale. We therefore emphasize the importance of balancing model capabilities with computational efficiency, as can be seen in traditional expert systems.

For practitioners, CareerBERT has the potential to significantly support career consultants in their daily work. By providing relevant job recommendations based on resumes, the model can help consultants identify suitable job matches more efficiently, ultimately improving the quality of their services \citep{lavi2021consultantbert}. This is particularly valuable when dealing with atypical career profiles or cross-domain transitions, where the system has demonstrated strong performance. However, the system's occasional difficulty in distinguishing between closely related roles with different skill requirements (such as Data Scientists versus Data Warehouse Developers) underscores its optimal use as a decision support tool within the consulting process rather than as a standalone solution.

Furthermore, our research highlights the potential of NLP-based tools in streamlining and enhancing various aspects of the job market, such as talent acquisition, career development, and workforce planning. Our evaluation results demonstrated the system's versatility across diverse career profiles, from traditional white-collar roles to blue-collar professions and atypical career paths, making it particularly valuable for career consultants who serve diverse client backgrounds. The successful application of CareerBERT in the domain of career consulting can inspire the development of similar tools in related fields, ultimately contributing to a more efficient and effective labor market. The reproducibility and multilingual potential of our approach, afforded by the use of freely available and language-agnostic data, further emphasizes the broad applicability and impact of our research in practice \citep{frankel2022disclosure}.

The potential impact of CareerBERT is significant, considering the urgent need for more effective, data-driven job matching and recommendation services in the rapidly evolving job market \citep{skills2014oecd, learning2010oecd}. With millions of people using platforms like LinkedIn\footnote{\href{https://news.linkedin.com/about-us\#Statistics}{https://news.linkedin.com/about-us\#Statistics}} to search for new jobs, the development of accurate and efficient job matching algorithms can lead to substantial economic benefits, such as reducing the global skills gap \citep{wef2023jobs} and increasing job satisfaction and performance \citep{cable2002convergent, soltis2023contextualizing, kristof2005consequences}.

\subsection{Limitations and Future Research Directions}
\label{subsec:limitations}

While our research has made significant contributions and demonstrated promising results, it is essential to acknowledge the limitations of our work and identify potential avenues for future research.

One limitation of our work is the reliance on job advertisements as a proxy for resumes in the application-grounded evaluation. Although this approach provided valuable insights into CareerBERT's performance, it may not fully capture the nuances and characteristics of real-world resumes. Future research could focus on collecting a larger dataset of resumes with annotated ESCO job codes to conduct a more direct evaluation of CareerBERT's performance.

A second limitation of our job recommendation system is the handling of low quality resumes and the cold start problem \citep{lika2014facing}. Our evaluation revealed that shorter resumes with limited information led to more diverse and less relevant recommendations across all models, indicating a systemic challenge. Additionally, we observed that individual keywords in short resumes can have a disproportionate influence on recommendations, potentially leading to misleading results. Resumes often vary in format, completeness, and relevance, with many lacking detailed or updated information, making accurate recommendations challenging. Additionally, the cold-start problem arises for users with limited prior experience, such as fresh graduates, or for emerging job titles with insufficient historical data, resulting in suboptimal matches. Addressing these issues requires more robust data enrichment techniques and mechanisms to better infer user intent from minimal input. A promising approach in this direction could be the incorporation of a prompt-based interactive resume completion method that leverages the generative capabilities of large language models \citep{du_enhancing_2024}. In this way, the quality of low-quality resumes could be enhanced with synthetically generated entries for improved job recommendations.


Another limitation of this study is the relatively small scale of the human-grounded evaluation, which involved five resumes and ten HR experts. While this evaluation provided valuable insights into the model's capabilities across different career profiles and demonstrated strong performance in most cases, further validation through methods such as A/B testing would be beneficial for assessing the system's performance at scale.
Although this evaluation provided valuable insights into the model's real-world applicability, future research could benefit from expanding the scale to include a larger and more diverse set of resumes and a broader panel of domain experts. However, it is important to note that our evaluation is quite extensive for HR experts, as each expert had to read through 143 recommendations and validate their fit and relevance to the resume at hand. Given the time-consuming nature of this task, consulting ten HR experts is considered sufficient for the purposes of this study. Nevertheless, we acknowledge that an expansion of the evaluation scale would allow for a more comprehensive assessment of CareerBERT's performance and its generalizability across different industries and job profiles \citep{lee2003generalizing}.

Furthermore, our research focused primarily on the German labor market and utilized German language models (GBERT and jobGBERT). While this targeted approach allowed us to demonstrate the effectiveness of our methodology in a specific linguistic context that may not be extensively covered by mainstream international research, it also limits the immediate generalizability of our findings to other languages and labor markets. However, the successful application of CareerBERT in the German language highlights the adaptability and potential of our approach to be readily transferred to other linguistic contexts  \citep{pires2019multilingual}. Future research could explore the performance of similar models in various cultural and linguistic settings, assessing the generalizability of our findings and the potential for developing comparable tools in international job markets. By extending the research to encompass a wider range of languages and labor markets, future studies could contribute to a more comprehensive understanding of the applicability and effectiveness of AI-driven career recommendation systems on a global scale.

Moreover, while our current architecture demonstrates a strong performance, future research should focus on developing specialized language models for job matching that combine the efficiency of expert systems with advanced semantic understanding. Our modular design specifically accommodates such future developments, positioning our work as a practical solution and a foundational contribution to the field. As shown in Table~\ref{tab:relatedwork}, we selected baselines where data availability and technical reproducibility (including model architectures, training procedures, and hyperparameters) were given to ensure a reliable implementation. Many approaches rely on proprietary resources or lack sufficient technical documentation, hindering a precise reconstruction. We plan to extend these comparisons if more complete implementations become available.

Another potential direction for future research is the integration of additional data sources and features into the CareerBERT model. For example, incorporating information about job seekers' skills, experiences, and preferences could enable more personalized and context-aware job recommendations. Exploring the integration of other data sources, such as social media profiles or educational backgrounds, could further enhance the accuracy and relevance of the recommendations. Nevertheless, the use of any data source should always be considered with caution because there are also ethical issues related to job recommendations based on personal data. This includes the potential for bias and discrimination if the system learns from historical data reflecting societal inequities, leading to unfair or stereotypical job recommendations. Misrepresentation or oversimplification of a candidate’s qualifications could result in mismatched opportunities, hindering career progress.

Lastly, future research could investigate the explainability and interpretability of the CareerBERT model. Developing methods to provide transparent and understandable explanations for the generated job recommendations could increase the trust and adoption of the model among career consultants and job seekers. Exploring techniques such as attention mechanisms or rule-based explanations could contribute to the development of more interpretable and user-friendly recommendation systems \citep{danilevsky2021explainability, rudin2019stop}.

In conclusion, our research on CareerBERT has made significant contributions to the field of NLP and job recommendation systems, demonstrating the effectiveness of domain adaptation and the potential of combining taxonomic data with real-world job advertisements. The implications of our work extend to both research and practice, offering valuable insights and inspiration for future developments in this domain. Despite the limitations of our study, the identified future research directions present exciting opportunities to further advance the field and develop more sophisticated and user-centric job recommendation tools.





\newpage
\appendix

\section{CareerBERT Application Interface}
\label{app:interface}

CareerBERT was deployed using Streamlit,\footnote{\href{https://streamlit.io/}{https://streamlit.io/}} a popular open-source framework for building interactive web applications for machine learning and data science projects. The Streamlit application provides a user-friendly interface that allows users to input their resumes and receive personalized job recommendations based on the CareerBERT model.

\begin{figure}[h!]
    \centering
    \includegraphics[width=\linewidth]{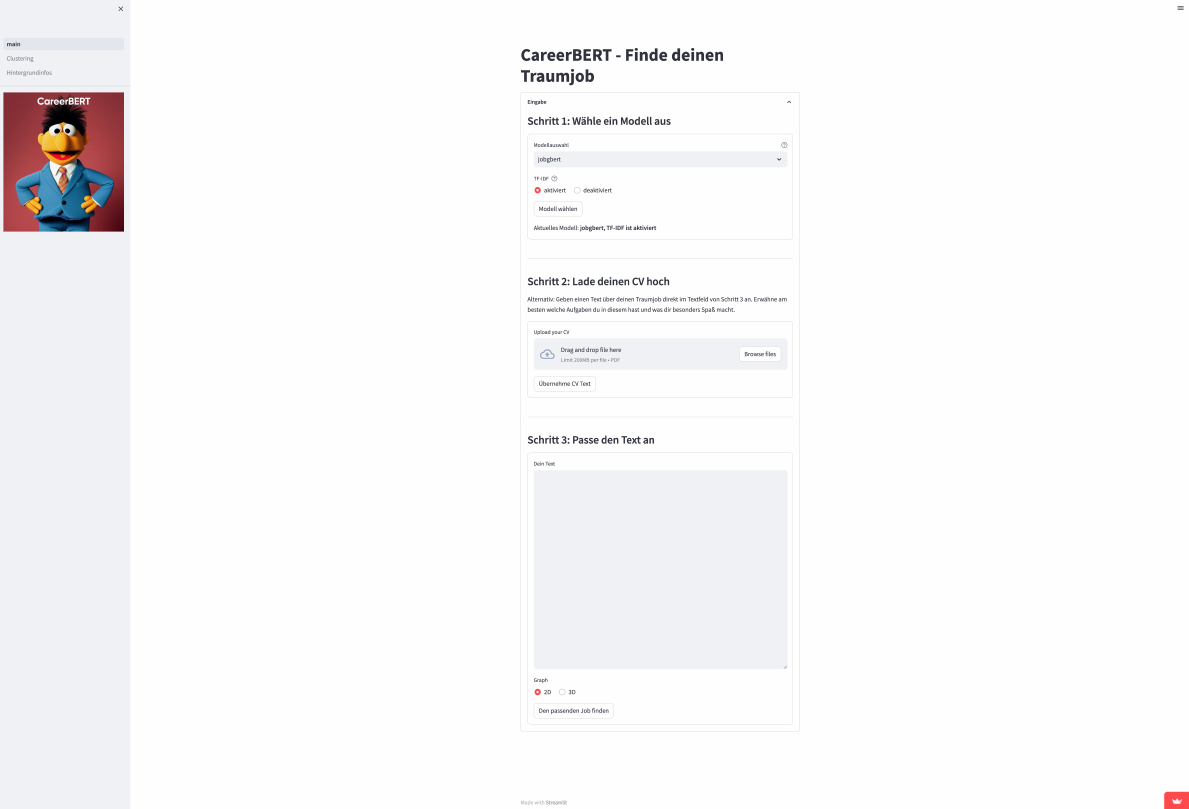}
    \caption{CareerBERT application interface deployed using Streamlit. Users can upload their resumes and receive personalized job recommendations based on the CareerBERT model. The interface allows users to select a model, upload a resume, and customize the text for better job matching.}
    \label{fig:application_interface}
\end{figure}

Figure \ref{fig:application_interface} showcases the CareerBERT application interface. Users can easily upload their resumes in PDF format, and the application processes the text and generates job recommendations. The interface displays the top job recommendations along with their corresponding ESCO codes and descriptions. Users can explore the recommended jobs and gain insights into how well their resumes match the suggested positions.

The CareerBERT application interface serves as a practical demonstration of how the CareerBERT model can be integrated into a real-world application to assist job seekers and career counselors in identifying suitable job opportunities based on individual resumes. The intuitive design and streamlined functionality of the interface make it accessible to a wide range of users, showcasing the potential of NLP-driven job recommendation systems in practice.

\section{Annotation Process and Classifier Training}
\label{app:classifier}
For annotating relevant paragraphs, we used the open-source tool Doccano.\footnote{\href{https://doccano.github.io/doccano/}{https://doccano.github.io/doccano/}} The annotation process was straightforward, as most job advertisements follow a consistent structure. We marked job titles, descriptions, and requirements as relevant, while considering all other paragraphs as irrelevant. Annotating 400 job advertisements resulted in a total of 2,444 paragraphs, with 1,395 marked as relevant and 1,049 as non-relevant. This balanced dataset was considered significant for training purposes and was further split into train (n = 1,833), validation (n = 306), and test (n = 305) sets.

To automatically classify the paragraphs into relevant and non-relevant, we trained a binary text classification model, consisting of a base model and a text classification head. As the base model, we used the jobGBERT model from \citet{gnehm2022evaluation}, to leverage its domain adaptation for both embedding creation and classifier training. As classification head, we used the default head for text classification from the HuggingFace transformers library\footnote{\href{https://huggingface.co/docs/transformers/tasks/sequence\_classification}{https://huggingface.co/docs/transformers/tasks/sequence\_classification}}, consisting of a linear layer with 768 input features and 768 output features, a dropout with a ratio of 0.1, and a linear layer with 768 input features and 1 output feature.
We fine-tuned the entire model (i.e., the base model and the text classification head) using the Trainer-class from the HuggingFace transformers library\footnote{\href{https://huggingface.co/docs/transformers/main\_classes/trainer}{https://huggingface.co/docs/transformers/main\_classes/trainer}}. In addition, we experimentally determined the optimal hyperparameter: batch size = 8, learning rate = 3e-5, and epochs = 10.
This configuration resulted in an accuracy of 0.96 and an F1-score of 0.97, demonstrating the effectiveness of our classifier in extracting relevant information from job advertisements.

 

\section{Description and Potential Challenges of Resumes}
\label{app:resumes}

All resumes are written in the German language. Due to space constraints and to maintain the anonymity of the individuals, we have included short descriptions of the five resumes in Table~\ref{tab:challenges} instead of providing the full text.

\begin{table}[h!]
  \small
  \centering
    \begin{tabular}{p{2cm}p{10cm}}
    \toprule
    Resume & Description \& Potential Challenges \\
    \midrule
    Resume 1  & This resume is focused on the medical domain, with the individual having experience as a nurse. Due to the homogeneity of the resume, it is expected that the models will primarily suggest jobs in the medical field. However, it may be challenging for the models to accurately identify that the person is not suitable for high-skilled medical positions, such as a physician, due to their educational background. \\
    Resume 2  & Resume 2 is notably short in length. While it is clear to a human reader that the person has relevant work experience in the sales domain, the limited text in the resume may pose challenges for the models. The hobbies section or section headlines (e.g., \textquote{Education}), which may not be as relevant, could potentially have a greater impact on the final recommendations. \\
    Resume 3  & This resume exhibits a clear focus on data analysis in recent years, with some influences from the HR domain. As a result, the expected recommendations are likely to reflect these two domains. \\
    Resume 4  & Resume 4 has a strong emphasis on the individual's experience as an industrial mechanic. This resume was chosen to include predictions for a blue-collar job and to assess the models' performance in this domain. \\
    Resume 5  & This resume stands out with its strong focus on environmental and political topics. It was selected due to its atypical field of politics, allowing for an exploration of recommendations in unconventional domains. \\
    \bottomrule
    \end{tabular}
    \caption{Descriptions of the utilized resumes and their potential challenges for job recommendation models.}
  \label{tab:challenges}
\end{table}

The decision to include short descriptions of the resumes, rather than the full text, was made for several reasons. First, providing the full text of the resumes would have significantly increased the length of the appendix, potentially making it unwieldy for readers. Second, and more importantly, including the full text of the resumes could have compromised the anonymity of the individuals whose resumes were used in the study. By providing concise descriptions that highlight the key characteristics and potential challenges associated with each resume, we strike a balance between providing sufficient information for readers to understand the nature of the resumes and protecting the privacy of the individuals involved.

These descriptions offer insights into the diverse range of experiences, skills, and domains represented in the resumes, as well as the potential challenges they may pose for job recommendation models. By including resumes from various fields, such as healthcare, sales, data analysis, industrial work, and politics, we aim to assess the performance and adaptability of the models across different domains and job types.

\section*{Acknowledgements}

\subsection*{Author Contributions}
\textbf{Julian Rosenberger:} Validation, Writing - Original Draft, Writing - Review \& Editing, Visualization, Project administration, 
\textbf{Lukas Wolfrum:} Conceptualization, Methodology, Software, Formal analysis, Investigation, Writing - Original Draft, 
\textbf{Sven Weinzierl:} Software, Formal analysis, Writing - Review \& Editing,  
\textbf{Mathias Kraus:} Validation, Writing - Review \& Editing,
\textbf{Patrick Zschech:} Conceptualization, Writing - Review \& Editing, Supervision

\subsection*{Funding Sources}
J.R., P.Z., and M.K. acknowledge funding from the Federal Ministry of Education and Research (BMBF) on ``White-box AI`` (Grant 01IS22080). P.Z. acknowledges funding from the Federal Ministry of Education and Research (BMBF) on ``AddIChron`` (Grant 16SV8995).

\subsection*{Declaration of Interests}
The authors have no competing interests to declare that are relevant to the content of this article.

\newpage
\bibliographystyle{elsarticle-harv} 
\bibliography{_references}

\begin{thebibliography}{51}
\expandafter\ifx\csname natexlab\endcsname\relax\def\natexlab#1{#1}\fi
\providecommand{\url}[1]{\texttt{#1}}
\providecommand{\href}[2]{#2}
\providecommand{\path}[1]{#1}
\providecommand{\DOIprefix}{doi:}
\providecommand{\ArXivprefix}{arXiv:}
\providecommand{\URLprefix}{URL: }
\providecommand{\Pubmedprefix}{pmid:}
\providecommand{\doi}[1]{\href{http://dx.doi.org/#1}{\path{#1}}}
\providecommand{\Pubmed}[1]{\href{pmid:#1}{\path{#1}}}
\providecommand{\bibinfo}[2]{#2}
\ifx\xfnm\relax \def\xfnm[#1]{\unskip,\space#1}\fi
\bibitem[{Autor(2015)}]{autor2015there}
\bibinfo{author}{Autor, D.H.}, \bibinfo{year}{2015}.
\newblock \bibinfo{title}{Why are there still so many jobs? the history and future of workplace automation}.
\newblock \bibinfo{journal}{Journal of Economic Perspectives} \bibinfo{volume}{29}, \bibinfo{pages}{3--30}.
\bibitem[{Bhatia et~al.(2019)Bhatia, Rawat, Kumar and Shah}]{bhatia2019end}
\bibinfo{author}{Bhatia, V.}, \bibinfo{author}{Rawat, P.}, \bibinfo{author}{Kumar, A.}, \bibinfo{author}{Shah, R.R.}, \bibinfo{year}{2019}.
\newblock \bibinfo{title}{End-to-end resume parsing and finding candidates for a job description using {BERT}}.
\newblock \bibinfo{journal}{arXiv preprint arXiv:1910.03089} .
\bibitem[{Borchert et~al.(2024)Borchert, Coussement, De~Weerdt and De~Caigny}]{borchert2024industry}
\bibinfo{author}{Borchert, P.}, \bibinfo{author}{Coussement, K.}, \bibinfo{author}{De~Weerdt, J.}, \bibinfo{author}{De~Caigny, A.}, \bibinfo{year}{2024}.
\newblock \bibinfo{title}{Industry-sensitive language modeling for business}.
\newblock \bibinfo{journal}{European Journal of Operational Research} \bibinfo{volume}{315}, \bibinfo{pages}{691--702}.
\bibitem[{Bressem et~al.(2024)Bressem, Papaioannou, Grundmann, Borchert, Adams, Liu, Busch, Xu, Loyen, Niehues et~al.}]{bressem2024medbert}
\bibinfo{author}{Bressem, K.K.}, \bibinfo{author}{Papaioannou, J.M.}, \bibinfo{author}{Grundmann, P.}, \bibinfo{author}{Borchert, F.}, \bibinfo{author}{Adams, L.C.}, \bibinfo{author}{Liu, L.}, \bibinfo{author}{Busch, F.}, \bibinfo{author}{Xu, L.}, \bibinfo{author}{Loyen, J.P.}, \bibinfo{author}{Niehues, S.M.}, et~al., \bibinfo{year}{2024}.
\newblock \bibinfo{title}{Medbert. de: A comprehensive german bert model for the medical domain}.
\newblock \bibinfo{journal}{Expert Systems with Applications} \bibinfo{volume}{237}, \bibinfo{pages}{121598}.
\bibitem[{Cable and DeRue(2002)}]{cable2002convergent}
\bibinfo{author}{Cable, D.M.}, \bibinfo{author}{DeRue, D.S.}, \bibinfo{year}{2002}.
\newblock \bibinfo{title}{The convergent and discriminant validity of subjective fit perceptions.}
\newblock \bibinfo{journal}{Journal of applied psychology} \bibinfo{volume}{87}, \bibinfo{pages}{875}.
\bibitem[{Campbell et~al.(2020)Campbell, Sands, Ferraro, Tsao and Mavrommatis}]{campbell2020data}
\bibinfo{author}{Campbell, C.}, \bibinfo{author}{Sands, S.}, \bibinfo{author}{Ferraro, C.}, \bibinfo{author}{Tsao, H.Y.J.}, \bibinfo{author}{Mavrommatis, A.}, \bibinfo{year}{2020}.
\newblock \bibinfo{title}{From data to action: How marketers can leverage ai}.
\newblock \bibinfo{journal}{Business horizons} \bibinfo{volume}{63}, \bibinfo{pages}{227--243}.
\bibitem[{Catelli et~al.(2022)Catelli, Fujita, De~Pietro and Esposito}]{catelli2022deceptive}
\bibinfo{author}{Catelli, R.}, \bibinfo{author}{Fujita, H.}, \bibinfo{author}{De~Pietro, G.}, \bibinfo{author}{Esposito, M.}, \bibinfo{year}{2022}.
\newblock \bibinfo{title}{Deceptive reviews and sentiment polarity: effective link by exploiting bert}.
\newblock \bibinfo{journal}{Expert Systems with Applications} \bibinfo{volume}{209}, \bibinfo{pages}{118290}.
\bibitem[{Chan et~al.(2020)Chan, Schweter and M{\"o}ller}]{chan2020gbert}
\bibinfo{author}{Chan, B.}, \bibinfo{author}{Schweter, S.}, \bibinfo{author}{M{\"o}ller, T.}, \bibinfo{year}{2020}.
\newblock \bibinfo{title}{{G}erman{'}s next language model}, in: \bibinfo{editor}{Scott, D.}, \bibinfo{editor}{Bel, N.}, \bibinfo{editor}{Zong, C.} (Eds.), \bibinfo{booktitle}{Proceedings of the 28th International Conference on Computational Linguistics}, pp. \bibinfo{pages}{6788--6796}.
\bibitem[{Chen et~al.(2024)Chen, Wei, Wang, Wang, Gong, Zhang and Miao}]{chen2024supplementing}
\bibinfo{author}{Chen, J.}, \bibinfo{author}{Wei, Z.}, \bibinfo{author}{Wang, J.}, \bibinfo{author}{Wang, R.}, \bibinfo{author}{Gong, C.}, \bibinfo{author}{Zhang, H.}, \bibinfo{author}{Miao, D.}, \bibinfo{year}{2024}.
\newblock \bibinfo{title}{Supplementing domain knowledge to bert with semi-structured information of documents}.
\newblock \bibinfo{journal}{Expert Systems with Applications} \bibinfo{volume}{235}, \bibinfo{pages}{121054}.
\bibitem[{Danilevsky et~al.(2021)Danilevsky, Dhanorkar, Li, Popa, Qian and Xu}]{danilevsky2021explainability}
\bibinfo{author}{Danilevsky, M.}, \bibinfo{author}{Dhanorkar, S.}, \bibinfo{author}{Li, Y.}, \bibinfo{author}{Popa, L.}, \bibinfo{author}{Qian, K.}, \bibinfo{author}{Xu, A.}, \bibinfo{year}{2021}.
\newblock \bibinfo{title}{Explainability for natural language processing}, in: \bibinfo{booktitle}{Proceedings of the 27th ACM SIGKDD Conference on Knowledge Discovery \& Data Mining}, pp. \bibinfo{pages}{4033--4034}.
\bibitem[{Dave et~al.(2018)Dave, Zhang, Al~Hasan, AlJadda and Korayem}]{dave_combined_2018}
\bibinfo{author}{Dave, V.S.}, \bibinfo{author}{Zhang, B.}, \bibinfo{author}{Al~Hasan, M.}, \bibinfo{author}{AlJadda, K.}, \bibinfo{author}{Korayem, M.}, \bibinfo{year}{2018}.
\newblock \bibinfo{title}{A {Combined} {Representation} {Learning} {Approach} for {Better} {Job} and {Skill} {Recommendation}}, in: \bibinfo{booktitle}{Proceedings of the 27th {ACM} {International} {Conference} on {Information} and {Knowledge} {Management}}, \bibinfo{publisher}{Association for Computing Machinery}, \bibinfo{address}{New York, NY, USA}. pp. \bibinfo{pages}{1997--2005}.
\bibitem[{Decorte et~al.(2021)Decorte, Van~Hautte, Demeester and Develder}]{decorte2021jobbert}
\bibinfo{author}{Decorte, J.J.}, \bibinfo{author}{Van~Hautte, J.}, \bibinfo{author}{Demeester, T.}, \bibinfo{author}{Develder, C.}, \bibinfo{year}{2021}.
\newblock \bibinfo{title}{{JobBERT}: Understanding job titles through skills}, in: \bibinfo{booktitle}{FEAST, ECML-PKDD 2021 Workshop, Proceedings}.
\bibitem[{Devlin et~al.(2018)Devlin, Chang, Lee and Toutanova}]{devlin2018bert}
\bibinfo{author}{Devlin, J.}, \bibinfo{author}{Chang, M.W.}, \bibinfo{author}{Lee, K.}, \bibinfo{author}{Toutanova, K.}, \bibinfo{year}{2018}.
\newblock \bibinfo{title}{Bert: Pre-training of deep bidirectional transformers for language understanding}.
\newblock \bibinfo{journal}{arXiv preprint arXiv:1810.04805} .
\bibitem[{Doshi-Velez and Kim(2017)}]{doshi2017towards}
\bibinfo{author}{Doshi-Velez, F.}, \bibinfo{author}{Kim, B.}, \bibinfo{year}{2017}.
\newblock \bibinfo{title}{Towards a rigorous science of interpretable machine learning}.
\newblock \bibinfo{journal}{arXiv preprint arXiv:1702.08608} .
\bibitem[{Du et~al.(2024)Du, Luo, Yan, Wang, Liu, Zhu, Song and Zhang}]{du_enhancing_2024}
\bibinfo{author}{Du, Y.}, \bibinfo{author}{Luo, D.}, \bibinfo{author}{Yan, R.}, \bibinfo{author}{Wang, X.}, \bibinfo{author}{Liu, H.}, \bibinfo{author}{Zhu, H.}, \bibinfo{author}{Song, Y.}, \bibinfo{author}{Zhang, J.}, \bibinfo{year}{2024}.
\newblock \bibinfo{title}{Enhancing {Job} {Recommendation} through {LLM}-{Based} {Generative} {Adversarial} {Networks}}.
\newblock \bibinfo{journal}{Proceedings of the AAAI Conference on Artificial Intelligence} \bibinfo{volume}{38}, \bibinfo{pages}{8363--8371}.
\bibitem[{Elsafty et~al.(2018)Elsafty, Riedl and Biemann}]{elsafty_document-based_2018}
\bibinfo{author}{Elsafty, A.}, \bibinfo{author}{Riedl, M.}, \bibinfo{author}{Biemann, C.}, \bibinfo{year}{2018}.
\newblock \bibinfo{title}{Document-based {Recommender} {System} for {Job} {Postings} using {Dense} {Representations}}, in: \bibinfo{editor}{Bangalore, S.}, \bibinfo{editor}{Chu-Carroll, J.}, \bibinfo{editor}{Li, Y.} (Eds.), \bibinfo{booktitle}{Proceedings of the 2018 {Conference} of the {North} {American} {Chapter} of the {Association} for {Computational} {Linguistics}: {Human} {Language} {Technologies}, {Volume} 3 ({Industry} {Papers})}, \bibinfo{publisher}{Association for Computational Linguistics}, \bibinfo{address}{New Orleans - Louisiana}. pp. \bibinfo{pages}{216--224}.
\bibitem[{Feuerriegel et~al.(2024)Feuerriegel, Hartmann, Janiesch and Zschech}]{feuerriegel2024generative}
\bibinfo{author}{Feuerriegel, S.}, \bibinfo{author}{Hartmann, J.}, \bibinfo{author}{Janiesch, C.}, \bibinfo{author}{Zschech, P.}, \bibinfo{year}{2024}.
\newblock \bibinfo{title}{Generative ai}.
\newblock \bibinfo{journal}{Business \& Information Systems Engineering} \bibinfo{volume}{66}, \bibinfo{pages}{111--126}.
\bibitem[{Frankel et~al.(2022)Frankel, Jennings and Lee}]{frankel2022disclosure}
\bibinfo{author}{Frankel, R.}, \bibinfo{author}{Jennings, J.}, \bibinfo{author}{Lee, J.}, \bibinfo{year}{2022}.
\newblock \bibinfo{title}{Disclosure sentiment: Machine learning vs. dictionary methods}.
\newblock \bibinfo{journal}{Management Science} \bibinfo{volume}{68}, \bibinfo{pages}{5514--5532}.
\bibitem[{Gnehm et~al.(2022)Gnehm, B{\"u}hlmann and Clematide}]{gnehm2022evaluation}
\bibinfo{author}{Gnehm, A.S.}, \bibinfo{author}{B{\"u}hlmann, E.}, \bibinfo{author}{Clematide, S.}, \bibinfo{year}{2022}.
\newblock \bibinfo{title}{Evaluation of transfer learning and domain adaptation for analyzing german-speaking job advertisements}, in: \bibinfo{booktitle}{Proceedings of the Thirteenth Language Resources and Evaluation Conference}, pp. \bibinfo{pages}{3892--3901}.
\bibitem[{Gururangan et~al.(2020)Gururangan, Marasović, Swayamdipta, Lo, Beltagy, Downey and Smith}]{gururangan2020don}
\bibinfo{author}{Gururangan, S.}, \bibinfo{author}{Marasović, A.}, \bibinfo{author}{Swayamdipta, S.}, \bibinfo{author}{Lo, K.}, \bibinfo{author}{Beltagy, I.}, \bibinfo{author}{Downey, D.}, \bibinfo{author}{Smith, N.A.}, \bibinfo{year}{2020}.
\newblock \bibinfo{title}{Don't {Stop} {Pretraining}: {Adapt} {Language} {Models} to {Domains} and {Tasks}}, in: \bibinfo{editor}{Jurafsky, D.}, \bibinfo{editor}{Chai, J.}, \bibinfo{editor}{Schluter, N.}, \bibinfo{editor}{Tetreault, J.} (Eds.), \bibinfo{booktitle}{Proceedings of the 58th {Annual} {Meeting} of the {Association} for {Computational} {Linguistics}}, \bibinfo{publisher}{Association for Computational Linguistics}, \bibinfo{address}{Online}. pp. \bibinfo{pages}{8342--8360}.
\bibitem[{Hardy et~al.(2022)Hardy, Tey, Cyrus-Lai, Martell, Olstad and Uhlmann}]{hardy2022bias}
\bibinfo{author}{Hardy, J.H.}, \bibinfo{author}{Tey, K.S.}, \bibinfo{author}{Cyrus-Lai, W.}, \bibinfo{author}{Martell, R.F.}, \bibinfo{author}{Olstad, A.}, \bibinfo{author}{Uhlmann, E.L.}, \bibinfo{year}{2022}.
\newblock \bibinfo{title}{Bias in context: Small biases in hiring evaluations have big consequences}.
\newblock \bibinfo{journal}{Journal of Management} \bibinfo{volume}{48}, \bibinfo{pages}{657--692}.
\bibitem[{Helaly et~al.(2022)Helaly, Rady and Aref}]{helaly2022bert}
\bibinfo{author}{Helaly, M.A.}, \bibinfo{author}{Rady, S.}, \bibinfo{author}{Aref, M.M.}, \bibinfo{year}{2022}.
\newblock \bibinfo{title}{Bert contextual embeddings for taxonomic classification of bacterial dna sequences}.
\newblock \bibinfo{journal}{Expert Systems with Applications} \bibinfo{volume}{208}, \bibinfo{pages}{117972}.
\bibitem[{Kristof-Brown et~al.(2005)Kristof-Brown, Zimmerman and Johnson}]{kristof2005consequences}
\bibinfo{author}{Kristof-Brown, A.L.}, \bibinfo{author}{Zimmerman, R.D.}, \bibinfo{author}{Johnson, E.C.}, \bibinfo{year}{2005}.
\newblock \bibinfo{title}{Consequences of individuals'fit at work: A meta-analysis of person--job, person--organization, person--group, and person--supervisor fit}.
\newblock \bibinfo{journal}{Personnel psychology} \bibinfo{volume}{58}, \bibinfo{pages}{281--342}.
\bibitem[{Lavi et~al.(2021)Lavi, Medentsiy and Graus}]{lavi2021consultantbert}
\bibinfo{author}{Lavi, D.}, \bibinfo{author}{Medentsiy, V.}, \bibinfo{author}{Graus, D.}, \bibinfo{year}{2021}.
\newblock \bibinfo{title}{{conSultantBERT}: {Fine}-tuned {Siamese} {Sentence}-{BERT} for {Matching} {Jobs} and {Job} {Seekers}}, in: \bibinfo{booktitle}{Proceedings of the {First} {Workshop} on {Recommender} {Systems} in {Human} {Resources} ({RecSys} in {HR} 2021)}, \bibinfo{address}{Amsterdam, Netherlands}.
\bibitem[{Lazaroiu et~al.(2024)Lazaroiu, Gedeon, Rogalska, Valaskova, Nagy, Musa, Zvarikova, Poliak, Horak, Crețoiu et~al.}]{lazaroiu2024digital}
\bibinfo{author}{Lazaroiu, G.}, \bibinfo{author}{Gedeon, T.}, \bibinfo{author}{Rogalska, E.}, \bibinfo{author}{Valaskova, K.}, \bibinfo{author}{Nagy, M.}, \bibinfo{author}{Musa, H.}, \bibinfo{author}{Zvarikova, K.}, \bibinfo{author}{Poliak, M.}, \bibinfo{author}{Horak, J.}, \bibinfo{author}{Crețoiu, R.I.}, et~al., \bibinfo{year}{2024}.
\newblock \bibinfo{title}{Digital twin-based cyber-physical manufacturing systems, extended reality metaverse enterprise and production management algorithms, and internet of things financial and labor market technologies in generative artificial intelligence economics}.
\newblock \bibinfo{journal}{Oeconomia Copernicana} \bibinfo{volume}{15}, \bibinfo{pages}{837--870}.
\bibitem[{Lazaroiu and Rogalska(2023)}]{lazaroiu2023generative}
\bibinfo{author}{Lazaroiu, G.}, \bibinfo{author}{Rogalska, E.}, \bibinfo{year}{2023}.
\newblock \bibinfo{title}{How generative artificial intelligence technologies shape partial job displacement and labor productivity growth}.
\newblock \bibinfo{journal}{Oeconomia Copernicana} \bibinfo{volume}{14}, \bibinfo{pages}{703--706}.
\bibitem[{Lee and Baskerville(2003)}]{lee2003generalizing}
\bibinfo{author}{Lee, A.S.}, \bibinfo{author}{Baskerville, R.L.}, \bibinfo{year}{2003}.
\newblock \bibinfo{title}{Generalizing generalizability in information systems research}.
\newblock \bibinfo{journal}{Information Systems Research} \bibinfo{volume}{14}, \bibinfo{pages}{221--243}.
\bibitem[{Li et~al.(2023)Li, Kang and De~Bie}]{li2023skillgpt}
\bibinfo{author}{Li, N.}, \bibinfo{author}{Kang, B.}, \bibinfo{author}{De~Bie, T.}, \bibinfo{year}{2023}.
\newblock \bibinfo{title}{Skillgpt: a restful api service for skill extraction and standardization using a large language model}.
\newblock \bibinfo{journal}{arXiv preprint arXiv:2304.11060} .
\bibitem[{Lika et~al.(2014)Lika, Kolomvatsos and Hadjiefthymiades}]{lika2014facing}
\bibinfo{author}{Lika, B.}, \bibinfo{author}{Kolomvatsos, K.}, \bibinfo{author}{Hadjiefthymiades, S.}, \bibinfo{year}{2014}.
\newblock \bibinfo{title}{Facing the cold start problem in recommender systems}.
\newblock \bibinfo{journal}{Expert systems with applications} \bibinfo{volume}{41}, \bibinfo{pages}{2065--2073}.
\bibitem[{Malik et~al.(2023)Malik, Budhwar, Mohan and NR}]{malik2023employee}
\bibinfo{author}{Malik, A.}, \bibinfo{author}{Budhwar, P.}, \bibinfo{author}{Mohan, H.}, \bibinfo{author}{NR, S.}, \bibinfo{year}{2023}.
\newblock \bibinfo{title}{Employee experience--the missing link for engaging employees: Insights from an mne's ai-based hr ecosystem}.
\newblock \bibinfo{journal}{Human Resource Management} \bibinfo{volume}{62}, \bibinfo{pages}{97--115}.
\bibitem[{Mikolov et~al.(2013)Mikolov, Chen, Corrado and Dean}]{mikolov2013efficient}
\bibinfo{author}{Mikolov, T.}, \bibinfo{author}{Chen, K.}, \bibinfo{author}{Corrado, G.}, \bibinfo{author}{Dean, J.}, \bibinfo{year}{2013}.
\newblock \bibinfo{title}{Efficient estimation of word representations in vector space}.
\newblock \bibinfo{journal}{arXiv preprint arXiv:1301.3781} .
\bibitem[{Musset and Kurekova(2018)}]{musset2018working}
\bibinfo{author}{Musset, P.}, \bibinfo{author}{Kurekova, L.M.}, \bibinfo{year}{2018}.
\newblock \bibinfo{title}{Working it out: Career guidance and employer engagement}.
\newblock \bibinfo{publisher}{OECD}.
\bibitem[{OECD(2010)}]{learning2010oecd}
\bibinfo{author}{OECD}, \bibinfo{year}{2010}.
\newblock \bibinfo{title}{Learning for Jobs}.
\newblock \bibinfo{publisher}{OECD Paris, France}.
\bibitem[{OECD(2014)}]{skills2014oecd}
\bibinfo{author}{OECD}, \bibinfo{year}{2014}.
\newblock \bibinfo{title}{Skills beyond School}.
\newblock \bibinfo{publisher}{OECD Paris, France}.
\bibitem[{OECD(2019)}]{organisation2019oecd}
\bibinfo{author}{OECD}, \bibinfo{year}{2019}.
\newblock \bibinfo{title}{OECD skills outlook 2019: Thriving in a digital world}.
\newblock \bibinfo{publisher}{OECD Paris, France}.
\bibitem[{Pennington et~al.(2014)Pennington, Socher and Manning}]{pennington2014glove}
\bibinfo{author}{Pennington, J.}, \bibinfo{author}{Socher, R.}, \bibinfo{author}{Manning, C.D.}, \bibinfo{year}{2014}.
\newblock \bibinfo{title}{Glove: Global vectors for word representation}, in: \bibinfo{booktitle}{Proceedings of the 2014 conference on empirical methods in natural language processing (EMNLP)}, pp. \bibinfo{pages}{1532--1543}.
\bibitem[{Pires et~al.(2019)Pires, Schlinger and Garrette}]{pires2019multilingual}
\bibinfo{author}{Pires, T.}, \bibinfo{author}{Schlinger, E.}, \bibinfo{author}{Garrette, D.}, \bibinfo{year}{2019}.
\newblock \bibinfo{title}{How {Multilingual} is {Multilingual} {BERT}?}, in: \bibinfo{editor}{Korhonen, A.}, \bibinfo{editor}{Traum, D.}, \bibinfo{editor}{Màrquez, L.} (Eds.), \bibinfo{booktitle}{Proceedings of the 57th {Annual} {Meeting} of the {Association} for {Computational} {Linguistics}}, \bibinfo{publisher}{Association for Computational Linguistics}, \bibinfo{address}{Florence, Italy}. pp. \bibinfo{pages}{4996--5001}.
\bibitem[{Ramos et~al.(2003)}]{ramos2003using}
\bibinfo{author}{Ramos, J.}, et~al., \bibinfo{year}{2003}.
\newblock \bibinfo{title}{Using tf-idf to determine word relevance in document queries}, in: \bibinfo{booktitle}{Proceedings of the first instructional conference on machine learning}, \bibinfo{organization}{Citeseer}. pp. \bibinfo{pages}{29--48}.
\bibitem[{Reimers and Gurevych(2019)}]{reimers2019sentence}
\bibinfo{author}{Reimers, N.}, \bibinfo{author}{Gurevych, I.}, \bibinfo{year}{2019}.
\newblock \bibinfo{title}{Sentence-{BERT}: {Sentence} {Embeddings} using {Siamese} {BERT}-{Networks}}, in: \bibinfo{editor}{Inui, K.}, \bibinfo{editor}{Jiang, J.}, \bibinfo{editor}{Ng, V.}, \bibinfo{editor}{Wan, X.} (Eds.), \bibinfo{booktitle}{Proceedings of the 2019 {Conference} on {Empirical} {Methods} in {Natural} {Language} {Processing} and the 9th {International} {Joint} {Conference} on {Natural} {Language} {Processing} ({EMNLP}-{IJCNLP})}, \bibinfo{publisher}{Association for Computational Linguistics}, \bibinfo{address}{Hong Kong, China}. pp. \bibinfo{pages}{3982--3992}.
\bibitem[{Rudin(2019)}]{rudin2019stop}
\bibinfo{author}{Rudin, C.}, \bibinfo{year}{2019}.
\newblock \bibinfo{title}{Stop explaining black box machine learning models for high stakes decisions and use interpretable models instead}.
\newblock \bibinfo{journal}{Nature machine intelligence} \bibinfo{volume}{1}, \bibinfo{pages}{206--215}.
\bibitem[{Schopf et~al.(2021)Schopf, Braun and Matthes}]{schopf2021lbl2vec}
\bibinfo{author}{Schopf, T.}, \bibinfo{author}{Braun, D.}, \bibinfo{author}{Matthes, F.}, \bibinfo{year}{2021}.
\newblock \bibinfo{title}{Lbl2vec: An embedding-based approach for unsupervised document retrieval on predefined topics}, in: \bibinfo{booktitle}{Proceedings of the 17th International Conference on Web Information Systems and Technologies - WEBIST}, pp. \bibinfo{pages}{124--132}.
\bibitem[{Siebers et~al.(2022)Siebers, Janiesch and Zschech}]{siebers2022survey}
\bibinfo{author}{Siebers, P.}, \bibinfo{author}{Janiesch, C.}, \bibinfo{author}{Zschech, P.}, \bibinfo{year}{2022}.
\newblock \bibinfo{title}{A survey of text representation methods and their genealogy}.
\newblock \bibinfo{journal}{IEEE Access} \bibinfo{volume}{10}, \bibinfo{pages}{96492--96513}.
\bibitem[{Soltis et~al.(2023)Soltis, Dineen and Wolfson}]{soltis2023contextualizing}
\bibinfo{author}{Soltis, S.M.}, \bibinfo{author}{Dineen, B.R.}, \bibinfo{author}{Wolfson, M.A.}, \bibinfo{year}{2023}.
\newblock \bibinfo{title}{Contextualizing social networks: The role of person--organization fit in the network--job performance relationship}.
\newblock \bibinfo{journal}{Human Resource Management} \bibinfo{volume}{62}, \bibinfo{pages}{445--460}.
\bibitem[{Teigland(2019)}]{teigland2019digital}
\bibinfo{author}{Teigland, R.}, \bibinfo{year}{2019}.
\newblock \bibinfo{title}{The Digital Transformation of Labor: Automation, the Gig Economy and Welfare}.
\newblock \bibinfo{publisher}{Routledge}.
\bibitem[{Wang et~al.(2021)Wang, Reimers and Gurevych}]{wang2021tsdae}
\bibinfo{author}{Wang, K.}, \bibinfo{author}{Reimers, N.}, \bibinfo{author}{Gurevych, I.}, \bibinfo{year}{2021}.
\newblock \bibinfo{title}{{TSDAE}: {Using} {Transformer}-based {Sequential} {Denoising} {Auto}-{Encoderfor} {Unsupervised} {Sentence} {Embedding} {Learning}}, in: \bibinfo{editor}{Moens, M.F.}, \bibinfo{editor}{Huang, X.}, \bibinfo{editor}{Specia, L.}, \bibinfo{editor}{Yih, S.W.t.} (Eds.), \bibinfo{booktitle}{Findings of the {Association} for {Computational} {Linguistics}: {EMNLP} 2021}, \bibinfo{publisher}{Association for Computational Linguistics}, \bibinfo{address}{Punta Cana, Dominican Republic}. pp. \bibinfo{pages}{671--688}.
\bibitem[{Webersinke et~al.(2021)Webersinke, Kraus, Bingler and Leippold}]{webersinke2021climatebert}
\bibinfo{author}{Webersinke, N.}, \bibinfo{author}{Kraus, M.}, \bibinfo{author}{Bingler, J.A.}, \bibinfo{author}{Leippold, M.}, \bibinfo{year}{2021}.
\newblock \bibinfo{title}{Climatebert: A pretrained language model for climate-related text}.
\newblock \bibinfo{journal}{arXiv preprint arXiv:2110.12010} .
\bibitem[{{World Economic Forum}(2023)}]{wef2023jobs}
\bibinfo{author}{{World Economic Forum}}, \bibinfo{year}{2023}.
\newblock \bibinfo{title}{Future of jobs report 2023}.
\newblock \bibinfo{howpublished}{\url{https://www3.weforum.org/docs/WEF_Future_of_Jobs_2023.pdf}}.
\bibitem[{Yang et~al.(2017)Yang, Korayem, AlJadda, Grainger and Natarajan}]{yang_combining_2017}
\bibinfo{author}{Yang, S.}, \bibinfo{author}{Korayem, M.}, \bibinfo{author}{AlJadda, K.}, \bibinfo{author}{Grainger, T.}, \bibinfo{author}{Natarajan, S.}, \bibinfo{year}{2017}.
\newblock \bibinfo{title}{Combining content-based and collaborative filtering for job recommendation system}.
\newblock \bibinfo{journal}{Know.-Based Syst.} \bibinfo{volume}{136}, \bibinfo{pages}{37--45}.
\bibitem[{Zhang et~al.(2023)Zhang, van~der Goot and Plank}]{zhang2023escoxlm}
\bibinfo{author}{Zhang, M.}, \bibinfo{author}{van~der Goot, R.}, \bibinfo{author}{Plank, B.}, \bibinfo{year}{2023}.
\newblock \bibinfo{title}{{ESCOXLM}-{R}: Multilingual taxonomy-driven pre-training for the job market domain}, in: \bibinfo{editor}{Rogers, A.}, \bibinfo{editor}{Boyd-Graber, J.}, \bibinfo{editor}{Okazaki, N.} (Eds.), \bibinfo{booktitle}{Proceedings of the 61st Annual Meeting of the Association for Computational Linguistics (Volume 1: Long Papers)}, \bibinfo{publisher}{Association for Computational Linguistics}, \bibinfo{address}{Toronto, Canada}. pp. \bibinfo{pages}{11871--11890}.
\bibitem[{Zhang and Gu(2010)}]{zhang2010using}
\bibinfo{author}{Zhang, S.}, \bibinfo{author}{Gu, M.}, \bibinfo{year}{2010}.
\newblock \bibinfo{title}{Using text categorization to find job opportunities}, in: \bibinfo{booktitle}{2010 International Conference on Web Information Systems and Mining}, \bibinfo{organization}{IEEE}. pp. \bibinfo{pages}{25--29}.
\bibitem[{Zhao et~al.(2021)Zhao, Wang, Sigdel, Zhang, Hoang, Liu and Korayem}]{zhao2021embedding}
\bibinfo{author}{Zhao, J.}, \bibinfo{author}{Wang, J.}, \bibinfo{author}{Sigdel, M.}, \bibinfo{author}{Zhang, B.}, \bibinfo{author}{Hoang, P.}, \bibinfo{author}{Liu, M.}, \bibinfo{author}{Korayem, M.}, \bibinfo{year}{2021}.
\newblock \bibinfo{title}{Embedding-based {Recommender} {System} for {Job} to {Candidate} {Matching} on {Scale}}, in: \bibinfo{booktitle}{Proceedings of {The} 27th {ACM} {SIGKDD} {Conference} on {Knowledge} {Discovery} and {Data} {Mining} ({KDD}’21 {IRS} {Workshop})}, \bibinfo{publisher}{ACM}, \bibinfo{address}{New York, NY, USA}.

\end{thebibliography}





\end{document}